\makeatletter
\DeclareRobustCommand\onedot{\futurelet\@let@token\@onedot}
\def\@onedot{\ifx\@let@token.\else.\null\fi\xspace}

\def\ie{\emph{i.e}\onedot}

\makeatother
\documentclass[10pt,journal,compsoc]{IEEEtran}

\usepackage[linesnumbered,lined,vlined,ruled,commentsnumbered]{algorithm2e}
\usepackage{multirow} 
\usepackage{graphicx}
\usepackage{booktabs}
\usepackage{amssymb}
\usepackage{multirow,xspace}
\usepackage{makecell}
\usepackage{amsmath, eucal}
\usepackage{times}
\usepackage{epsfig}
\usepackage{graphicx}
\usepackage{float}
\floatstyle{plaintop}
\restylefloat{table}
\usepackage{caption}
\usepackage{amsmath}
\usepackage{amssymb}
\usepackage{bm}
\usepackage{dsfont}
\usepackage{subcaption}
\usepackage{blindtext}
\usepackage{colortbl,booktabs}
\usepackage{multirow}
\usepackage{booktabs}
\usepackage[english]{babel}
\usepackage[utf8x]{inputenc}
\usepackage{algpseudocode} 
\newcommand{\minus}{\scalebox{0.75}[1.0]{$-$}}
\usepackage[pagebackref=true,breaklinks=true,letterpaper=true,colorlinks,bookmarks=false]{hyperref}

\ifCLASSOPTIONcompsoc
  \usepackage[nocompress]{cite}
\else
  \usepackage{cite}
\fi

\usepackage{caption}
\usepackage[margin=4pt,font=footnotesize,labelfont=bf,labelsep=endash,tableposition=top]{caption}

\usepackage{subcaption}

\ifCLASSINFOpdf
\else
\fi

\begin{document}
\title{Open Set Recognition with Conditional Probabilistic Generative Models}

\author{Xin~Sun$^*$\thanks{$^*$ indicates equal contribution.},
        Chi~Zhang$^*$,
        Guosheng~Lin,
        and~Keck-Voon LING%
\IEEEcompsocitemizethanks{
\IEEEcompsocthanksitem X. Sun is with the School of Electrical and Electronic Engineering, Nanyang Technological University, Singapore 639798. \protect\\
E-mail: xin001@e.ntu.edu.sg.

\IEEEcompsocthanksitem C. Zhang is with the School of Computer Science and Engineering, Nanyang Technological University, Singapore 639798. \protect\\
E-mail: chi007@e.ntu.edu.sg.

\IEEEcompsocthanksitem G. Lin is with the School of Computer Science and Engineering, Nanyang Technological University, Singapore 639798. \protect\\
E-mail: gslin@ntu.edu.sg.

\IEEEcompsocthanksitem KV. LING is with the School of Electrical and Electronic Engineering, Nanyang Technological University, Singapore 639798. \protect\\
E-mail: ekvling@ntu.edu.sg.

}%
\thanks{Corresponding author: Guosheng Lin}

}

\IEEEtitleabstractindextext{%

\begin{abstract}

 Deep neural networks have made breakthroughs in a wide range of visual understanding tasks.
 A typical challenge that hinders their real-world applications is that unknown samples may be fed into the system during the testing phase, but traditional deep neural networks will wrongly recognize these unknown samples as one of the known classes. Open set recognition (OSR) is a potential solution to overcome this problem, where the open set classifier should have the flexibility to reject unknown samples and meanwhile maintain high classification accuracy in known classes. Probabilistic generative models, such as Variational Autoencoders (VAE) and Adversarial Autoencoders (AAE), are popular methods to detect unknowns, but they cannot provide discriminative representations for known classification. 
 In this paper, we propose a novel framework, called Conditional Probabilistic Generative Models (CPGM), for open set recognition. 
 The core insight of our work is to add discriminative information into the probabilistic generative models, such that the proposed models can not only detect unknown samples but also classify known classes by forcing different latent features to approximate conditional Gaussian distributions. We discuss many model variants and provide comprehensive experiments to study their characteristics.
 Experiment results on multiple benchmark datasets reveal that the proposed method significantly outperforms the baselines and achieves new state-of-the-art performance. 

\end{abstract}

}

\maketitle

\IEEEdisplaynontitleabstractindextext

\IEEEpeerreviewmaketitle

\IEEEraisesectionheading{\section{Introduction}\label{sec:introduction}}
\begin{figure}
\begin{subfigure}{0.48\columnwidth}
  \centering
  \includegraphics[height=3.2cm,width=3.5cm]{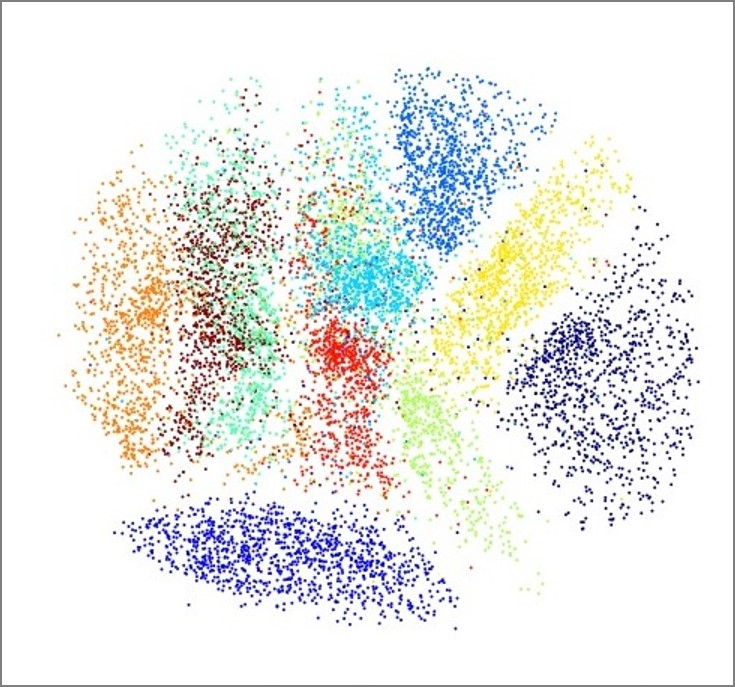}
  \caption{VAE}
  \label{fig:vae}
\end{subfigure}%
\hspace{.2in}
\begin{subfigure}{0.48\columnwidth}
  \centering
  \includegraphics[height=3.2cm,width=3.5cm]{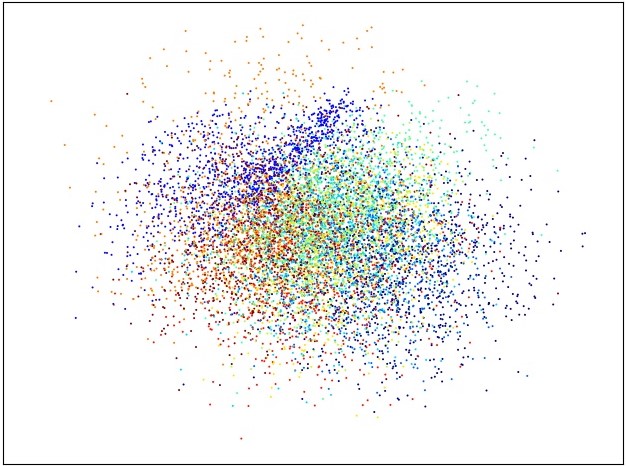}
  \caption{AAE}
  \label{fig:aae}
\end{subfigure}
\hspace{.2in}
\begin{subfigure}{0.48\columnwidth}
  \centering
  \includegraphics[height=3.2cm,width=3.5cm]{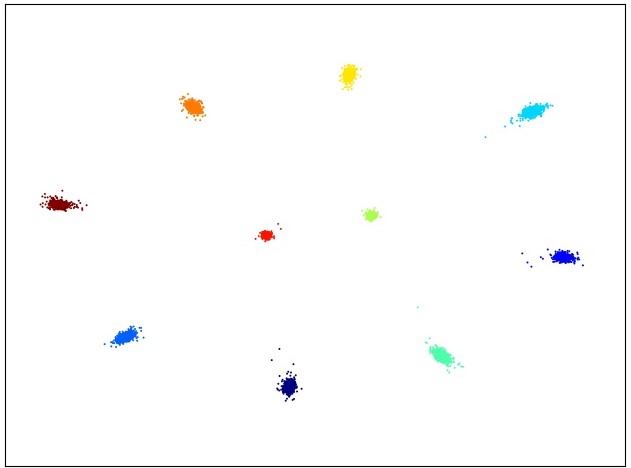}
  \caption{Ours: CPGM-VAE}
  \label{fig:cvae}
\end{subfigure}%
\hspace{.2in}
\begin{subfigure}{0.48\columnwidth}
  \centering
  \includegraphics[height=3.2cm,width=3.5cm]{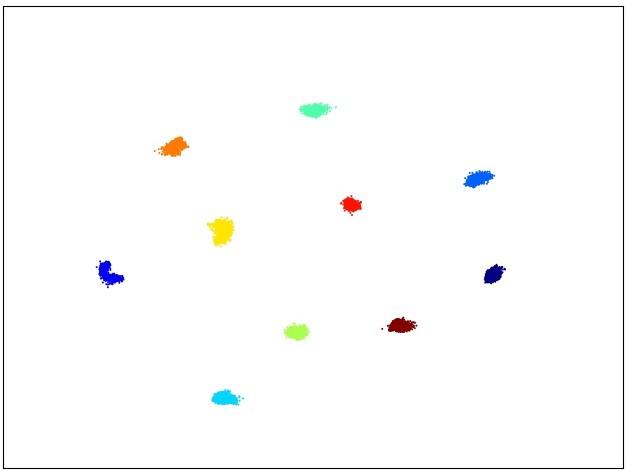}
  \caption{Ours: CPGM-AAE}
  \label{fig:caae}
\end{subfigure}
\hspace{.2in}
\caption{Comparison of latent representations on the MNIST dataset of VAE \textbf{(a)}, AAE \textbf{(b)}, and proposed methods: CPGM-VAE \textbf{(c)} and CPGM-AAE \textbf{(d)}. VAE and AAE are widely used in unknown detection, but they cannot provide discriminative features to undertake classification tasks as all features just follow one distribution. Comparatively, the proposed method can learn conditional Gaussian distributions by forcing different latent features to approximate different Gaussian models, which enables the proposed method to classify known samples as well as reject unknown samples.}
\end{figure}

\IEEEPARstart{I}{n} the past few years, deep learning has achieved remarkable success in many visual understanding tasks~\cite{mr-cnn,res-net,ssd,r-cnn}. However, there are still various challenges when they are applied to real-world scenarios. One typical challenge is that samples from unknown classes may be fed into the system during the testing phase. As the standard classification tasks are under a common closed set assumption: all training and testing data come from the same label space, these unknown samples will be recognized as one of the known classes. 

Open set recognition (OSR) \cite{toward} was proposed as a promising research direction to address this issue, where testing samples can come from any classes, including those that have not been seen during the training phase. Therefore, an open set classifier should have a dual character: unknown detection and known classification$\footnote{We refer to detection of unknown samples as \emph{unknown detection}, and classification of known samples as \emph{known classification}.}$. OSR is a challenging problem setting as only data from known classes are provided at training time, which prevents us from training an unknown sample detector in a fully supervised manner. To realize unknown detection, many previous works in the literature model the data distribution of known samples by unsupervised learning methods\cite{ad_cluster,deepsvdd,ad_aec,gmm, gpaae}, and those deviated data points from learned distribution are then deemed to be unknown samples. For example, Variational Autoencoder (VAE) \cite{vae} is a popular method and is often combined with  clustering \cite{ad_cluster}, GMM \cite{gmm}, or one-class \cite{deepsvdd} algorithms, to achieve this goal. VAE is a probabilistic graphical model which is trained not only to reconstruct the input accurately, but also to force the posterior distribution $q_{\bm{\phi}}(\bm{z}|\bm{x})$ in the latent space to approximate one prior distribution $p_{\bm{\theta}}(\bm{z})$, such as a multivariate Gaussian or Bernoulli distribution. The well-trained VAE is able to correctly describe the distribution of known data, and deviated samples which lie in low-probability regions will be assumed as unknown. In the same spirit, another line of works relies on deep generative adversarial networks (GAN) that were proposed, such as Adversarial Autoencoders (AAE) \cite{aae}, to perform variational inference and impose one prior distribution to the data distribution through adversarial training. AAE is able to match the posterior distribution in the latent space with one known prior distribution, such as a multivariate Gaussian distribution. Therefore, an AAE model is commonly trained through unsupervised learning or semi-supervised learning and widely applied to outlier detection \cite{spe_aae,ad_aae,ano_aae,video_aae}. Fig.~\ref{fig:vae} and Fig.~\ref{fig:aae} provide the visualization of the latent representations of VAE and AAE respectively in low-dimensional space when the prior distribution $p_{\bm{\theta}}(\bm{z})$ is a multivariate normal Gaussian. As we can see, although these probabilistic models excel at modeling data distributions of known samples for unknown detection, it cannot provide discriminative representations to undertake classification tasks as all data follow a universal distribution.

To overcome this shortcoming, in this paper, we propose a novel framework, Conditional Probabilistic Generative Models (CPGM), for open set recognition.
Different from previous anomaly detection methods based on probabilistic generative models, the proposed framework is able to generate class conditional posterior distributions $q_{\bm{\phi}}(\bm{z}|\bm{x},k)$ in the latent space where $k$ is the index of known classes. These conditional distributions are forced to approximate different multivariate Gaussian models $p_{\bm{\theta}}^{(k)}(\bm{z})={\mathcal{N}}(\bm{z};\bm{\mu}_{k},\textbf{I})$ where $\bm{\mu}_k$ is the mean of the $k$-th multivariate Gaussian distribution, obtained by a fully-connected layer that maps the categorical distribution of input to the latent space. Fig.~\ref{fig:cvae} and Fig.~\ref{fig:caae} are examples of latent representations of the proposed method on the MNIST dataset. These learned features will be fed to an open set classifier, which consists of two parts: an unknown detector and a closed set classifier. As known samples tend to follow prior distributions, the unknown detector will recognize those samples locating in lower probability regions as unknown. Meanwhile, for the known sample, the closed set classifier will calculate its prediction scores over all known classes and predict it as the class with the highest score.

In this paper, we mainly focus on two popular probabilistic generative models, namely VAE and AAE. To fully investigate their potentials for the goal of open-set classification, we explore many model variants and conduct comprehensive experiments to study their characteristics. For example, current networks for classification tend to have deep structures for higher accuracy \cite{deeper}, but traditional VAE models with deep hierarchies are difficult to optimize,  due to multiple layers of dependent stochastic variables \cite{ladder}.
We overcome this shortcoming by adopting the probabilistic ladder network \cite{ladder} in the proposed conditional VAE model, which enables the proposed VAE model to generate an approximate posterior distribution merging information from the bottom-up computed approximate likelihood with top-down prior information.

Through our comprehensive analysis in the experiment, we empirically demonstrate that our framework, with either VAE or AAE as the probabilistic generative model, can significantly outperform baseline methods and achieve new state-of-the-art performance. Our main contributions are summarized as follows:
\begin{itemize}
\setlength
  \item We propose a novel framework, Conditional Probabilistic Generative Models (CPGM), for open set recognition. Compared with the original VAE and AAE, the proposed method is able to learn conditional distributions for known classification and unknown detection.
  \item We explore many model variants of the proposed framework and present much practical guidance to use VAE and AAE for open set recognition.
 \item Experiments on several benchmark datasets show that our method outperforms existing methods and achieves new state-of-the-art performance.
\end{itemize}

\section{Related Work}
\textbf{Open Set Recognition (OSR).}  The approaches for OSR can be roughly divided into two branches: traditional methods (e.g., SVM \cite{toward,toward-pi}, sparse representation \cite{sparse}, Nearest Neighbor \cite{nearest}, etc.) and deep learning-based methods \cite{openmax,g-openmax,doc,counter,crosr,c2ae}. In traditional methods, Scheirer~\emph{et~al.}~\cite{toward} proposed an SVM based method which adds an extra hyper-line to detect unknown samples. Jain~\emph{et~al.}~\cite{toward-pi} proposed the $P_I$-SVM algorithm, which is able to reject unknown samples by adopting EVT to model the positive training samples at the decision boundary. Cevikalp~\emph{et~al.}~\cite{fafr,poly} defined the acceptance regions for known samples with a family of quasi-linear `polyhedral conic' functions. Zhang~\emph{et~al.}~\cite{sparse} pointed out that discriminative information is mostly hidden in the reconstruction error distributions, and proposed the sparse representation-based OSR model, called SROSR. Bendale~\emph{et~al.}~\cite{openworld} recognized unknown samples based on the distance between the testing samples and the centroids of the known classes. J{\'u}nior~\emph{et~al.}~\cite{nearest} proposed the Nearest Neighbor Distance Ratio (NNDR) technique, which carries out OSR according to the similarity score between the two most similar classes. 

Considering deep learning achieves state-of-the-art performance in a wide range of visual understanding tasks, deep learning-based OSR methods are gaining more and more attention. For example, Bendale~\emph{et~al.}~\cite{openmax} proposed the Openmax function to replace the Softmax function in CNNs. In this method, the probability distribution of Softmax is redistributed to get the class probability of unknown samples. Based on Openmax, Ge~\emph{et~al.}~\cite{g-openmax} proposed the Generative Openmax method, using generative models to synthesize unknown samples to train the network. Shu~\emph{et~al.}~\cite{doc} proposed the Deep Open Classifier (DOC) model, which replaces the Softmax layer with a 1-vs-rest layer containing sigmoid functions. Counterfactual image generation, a dataset augmentation technique proposed by Neal~\emph{et~al.}~\cite{counter}, aims to synthesize unknown-class images. Then the decision boundaries between unknown and known classes can be converged from these known-like but actually unknown sample sets. Yoshihashi~\emph{et~al.}~\cite{crosr} proposed the CROSR model, which combines the supervised learned prediction and unsupervised reconstructive latent representation to redistribute the probability distribution. Oza and Patel~\cite{c2ae} trained a class conditional auto-encoder (C2AE) to get the decision boundary from the reconstruction errors by extreme value theory (EVT). The training phase of C2AE is divided into two steps (closed-set training and open-set training), and a batch of samples need to be selected from training data to generate non-match reconstruction errors. This is difficult in practice and testing results are highly dependent on the selected samples. On the contrary, the proposed framework is an end-to-end system and does not need extra data pre-processing.

\textbf{Anomaly Detection.} Anomaly detection (also called outlier detection or unknown detection) aims to distinguish anomalous samples from normal samples, which can be introduced into OSR for unknown detection.  Some general anomaly detection methods are based on Support Vector Machine (SVM) \cite{svdd,ocsvm} or forests \cite{forest}. In recent years, deep neural networks have also been used in anomaly detection, mainly based on auto-encoders trained in an unsupervised manner \cite{ad_aec, ad_cluster, deepsvdd, gmm}. Auto-encoders commonly have a bottleneck architecture to induce the network to learn abstract latent representations. Meanwhile, these networks are typically trained by minimizing reconstruction errors. In anomaly detection, the training samples commonly come from the same distribution, thus the well-trained auto-encoders could extract the common latent representations from the normal samples and reconstruct them correctly, while anomalous samples do not contain these common latent representations and could not be reconstructed correctly. Although VAEs are widely applied in anomaly detection, it cannot provide discriminative features for classification tasks.

Apart from auto-encoders, some studies used Generative Adversarial Networks (GANs) to detect anomalies \cite{ad_gan,gpaae}. GANs are trained to generate similar samples according to the training samples. Given a testing sample, the GAN tries to find the point in the generator's latent space that can generate a sample closest to the input. Intuitively, the well-trained GAN could give good representations for normal samples and terrible representations for anomalies.

There are also some related topics that focus on handling novel classes. For example, incremental learning~\cite{liu2020mnemonics} aims to make predictions on both old classes and new classes without accessing data in old classes.
Few-shot learning~\cite{Zhang_2020_CVPR,zhang2019pyramid,zhang2019canet,liu2020crnet} aims to utilize a network trained on old classes to undertake vision tasks on unseen classes.
\section{Preliminary}
Before introducing the proposed method, we briefly introduce the terminology and notations of VAE \cite{vae} and AAE \cite{aae}. 
\subsection{VAE}
The VAE commonly consists of an encoder, a decoder and a loss function ${\mathcal{L}}(\bm{\theta};\bm{\phi};\bm{x})$. The encoder is a neural network that has parameters $\bm{\phi}$. Its input is a sample $\bm{x}$ and its output is a hidden representation $\bm{z}$. The decoder is another neural network with parameters $\theta$. Its input is the representation $\bm{z}$ and it outputs the probability distribution of the sample. The loss function in the VAE is defined as follows:

\begin{equation}
\begin{split}
{\mathcal{L}}(\bm{\theta};\bm{\phi};\bm{x})= & \ \minus D_{KL}\big(q_{\bm{\phi}}(\bm{z}|\bm{x}) \ || \ p_{\bm{\theta}}(\bm{z})\big)\\
& +\mathds{E}_{q_{\bm{\phi}}(\bm{z}|\bm{x})}\big[\log{p_{\bm{\theta}}(\bm{x}|\bm{z})}\big]
\label{loss_vae}
\end{split}
\end{equation}
where $q_{\bm{\phi}}(\bm{z}|\bm{x})$ is the approximate posterior, $p_{\bm{\theta}}(\bm{z})$ is the prior distribution of the latent representation $\bm{z}$ and $p_{\bm{\theta}}(\bm{x}|\bm{z})$ is the likelihood of the input $\bm{x}$ given latent representation $\bm{z}$. On the right-hand side of Eqn.~\ref{loss_vae}, the first term is the KL-divergence between the approximate posterior and the prior. It can be viewed as a regularizer to encounter the approximate posterior to be close to the prior $p_{\bm{\theta}}(\bm{z})$. The second term can be viewed as the reconstruction errors to force the output to reconstruct the input correctly.

Commonly, the prior over the latent representation $\bm{z}$ is the centered isotropic multivariate Gaussian $p_{\bm{\theta}}(\bm{z})={\mathcal{N}}(\bm{z};\bm{0},\textbf{I})$. In this case, the variational approximate posterior could be a multivariate Gaussian with a diagonal covariance structure:

\begin{equation}
q_{\bm{\phi}}(\bm{z}|\bm{x})={\mathcal{N}}(\bm{z};\bm{\mu},\bm{\sigma}^2\textbf{I})
\end{equation}
where the mean $\bm{\mu}$ and the standard deviation $\bm{\sigma}$ of the approximate posterior are outputs of the encoding multi-layered perceptrons (MLPs). The latent representation $\bm{z}$ is defined as $\bm{z}=\bm{\mu}+\bm{\sigma}\odot \bm{\epsilon}$ where $\bm{\epsilon}\sim {\mathcal{N}}(\bm{0},\textbf{I})$ and $\odot$ is the element-wise product. Let $J$ be the dimensionality of $\bm{z}$, then the KL-divergence can be calculated \cite{vae}:

\begin{equation}
\begin{aligned}
&\minus D_{KL}\big(q_{\bm{\phi}}(\bm{z}|\bm{x}\big) \ || \ p_{\bm{\theta}}(\bm{z}))\\
&=\frac{1}{2}\sum_{j=1}^{J}\big(1+\log(\sigma_{j}^{2})-\mu_{j}^{2}-\sigma_{j}^{2}\big)
\label{kl_vae}
\end{aligned}
\end{equation}

With the loss function ${\mathcal{L}}(\bm{\theta};\bm{\phi};\bm{x})$, the VAE is trained not only to reconstruct the input accurately, but also to force the posterior distribution $q_{\bm{\phi}}(\bm{z}|\bm{x})$ in the latent space  to approximate the prior distribution $p_{\bm{\theta}}(\bm{z})$. If a sample locates in the low probability region of the learned distribution, this sample will be recognized as unknown.

\subsection{AAE}
The AAE \cite{aae} model is composed of three modules: an encoder (generator) $G(\bm{x})$, a decoder $De(\bm{z})$ and a discriminator $D(p(\bm{z}))$. The AAE model is trained in two phases: the \emph{reconstruction} phase and the \emph{regularization} phase. In the \emph{reconstruction} phase, the encoder $G(\bm{x})$ and decoder $De(\bm{z})$ are trained jointly to reconstruct the input $\bm{x}$ correctly, which can be expressed as follows:
\begin{equation}
{\mathcal{L}_r}=||\bm{x}-\bm{\tilde x}||^2_2
\end{equation}
where $\bm{\tilde x}$ is the reconstructed image.

In the \emph{regularization} phase, an adversarial game is established between the encoder $G(\bm{x})$ and the discriminator $D(p(\bm{z}))$: firstly the discriminator $D(p(\bm{z}))$ is trained to distinguish the true distribution (generated using the prior distribution $p(\bm{z})$) from the posterior distribution $q(\bm{z}|\bm{x})$ (generated by the encoder $G(\bm{x})$), then the encoder $G(\bm{x})$ is trained to confuse the discriminator $D(p(\bm{z}))$. This adversarial process can be expressed in the following equation:

\begin{equation}
\mathop{min}\limits_{G}\mathop{max}\limits_{D}E_{p(\bm{z})}[logD(p(\bm{z}))]+E_{\bm{x}\sim p_{data}}[log(1-D(G(\bm{x})))]
\end{equation}

When training is completed, the posterior distribution $q(\bm{z}|\bm{x})$ in the latent space will be close to the prior distribution $p(\bm{z})$. If a sample locates in the low probability region of the learned distribution, this sample will be recognized as unknown.

\section{Method} 

In this section, we describe the proposed framework based on two probabilistic generative models, \ie, VAE and AAE, which are denoted by CPGM-VAE and CPGM-AAE, respectively. For each method, we first present the architecture of the proposed model and then describe the functions of different modules during training and testing.

\subsection{CPGM-VAE}
\subsubsection{Architecture}
\begin{figure} [htbp]
    \centering
    \includegraphics[width=0.95\columnwidth]{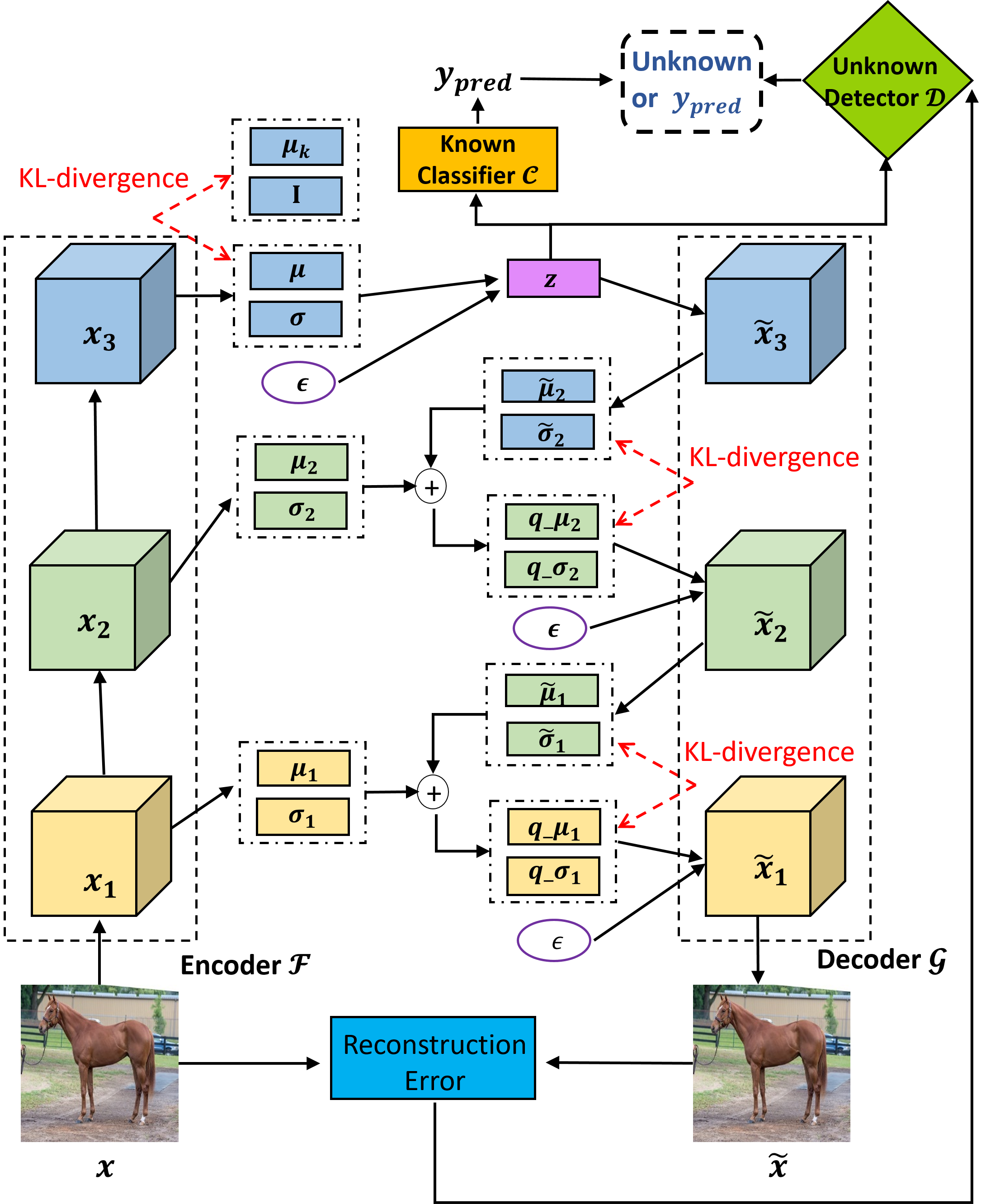}
    \caption{Block diagram of CPGM-VAE: The \textbf{encoder} $\bm{{\mathcal{F}}}$ and \textbf{decoder} $\bm{{\mathcal{G}}}$ are applied with the probabilistic ladder architecture to generate more informative posterior distributions. The \textbf{known classifier} $\bm{{\mathcal{C}}}$ takes latent representations as input and produces the probability distribution over the known classes. The \textbf{unknown detector} $\bm{{\mathcal{D}}}$ is modeled by the conditional Gaussian distributions and reconstruction errors from training samples, which is used for unknown detection. During training, the proposed model is trained to minimize the sum of the reconstruction loss ${\mathcal{L}_r}$, KL-divergence ${\mathcal{L}_{KL}}$ (both in the latent space and middle layers) and classification loss ${\mathcal{L}_c}$. During testing, the \textbf{unknown detector} $\bm{{\mathcal{D}}}$ will judge whether this sampler is unknown by its latent features and reconstruction errors. If this sample is known, the \textbf{known classifier} $\bm{{\mathcal{C}}}$ will give out its predicted label.}  
    \label{fig:framework}
\end{figure}

The architecture of CPGM-VAE is composed of four modules (as shown in Fig.~\ref{fig:framework}):\\
1. Encoder ${\mathcal{F}}$\\
2. Decoder ${\mathcal{G}}$\\
3. Known Classifier ${\mathcal{C}}$\\
4. Unknown Detector ${\mathcal{D}}$

\begin{figure} [htbp]
    \centering
    \includegraphics[width=\columnwidth]{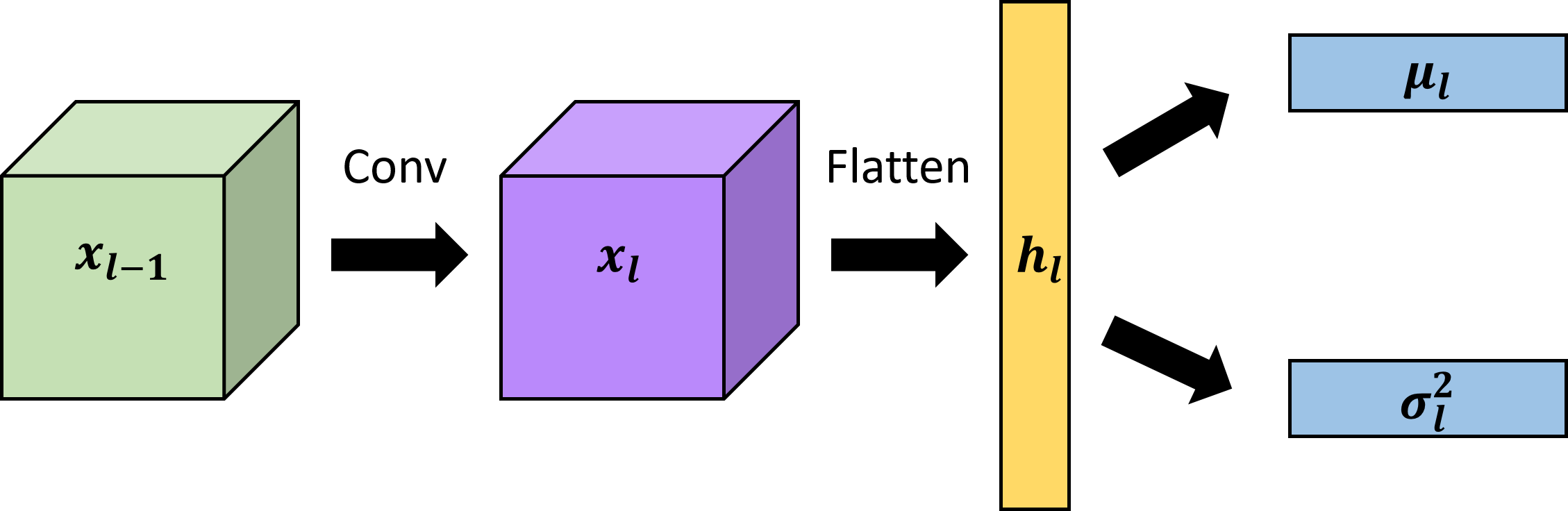}
    \caption{Operations in the upward pathway of CPGM-VAE.}
    \label{fig:o1}
\end{figure}

\textbf{Encoder} $\bm{{\mathcal{F}}.}$ To generate more informative posterior distributions, the probabilistic ladder architecture is adopted between the upward path and downward path. In detail, the $l$-th layer in the encoder ${\mathcal{F}}$ is expressed as follows:

\begin{equation*}
\begin{aligned}
&\bm{x}_l=\text{Conv}(\bm{x}_{l-1})\\
&\bm{h}_l=\text{Flatten}(\bm{x}_l)\\
&\bm{\mu}_l=\text{Linear}(\bm{h}_l)\\
&\bm{\sigma}^{2}_l=\text{Softplus}(\text{Linear}(\bm{h}_l))
\end{aligned}
\end{equation*} 
where \verb Conv  is a convolutional layer followed by a batch-norm layer and a PReLU layer, \verb Flatten  is a linear layer to flatten 2-dimensional data into 1-dimension, \verb Linear  is a single linear layer and \verb Softplus  applies $\log(1+\text{exp}(\cdotp))$ non-linearity to each component of its argument vector (Fig.~\ref{fig:o1} illustrates these operations). The latent representation $\bm{z}$ is defined as $\bm{z}=\bm{\mu}+\bm{\sigma}\odot \bm{\epsilon}$ where $\bm{\epsilon}\sim {\mathcal{N}}(\bm{0},\textbf{I})$, $\odot$ is the element-wise product, and $\bm{\mu}$, $\bm{\sigma}$ are the outputs of the top layer $L$.

\begin{figure} [htbp]
    \centering
    \includegraphics[width=\columnwidth]{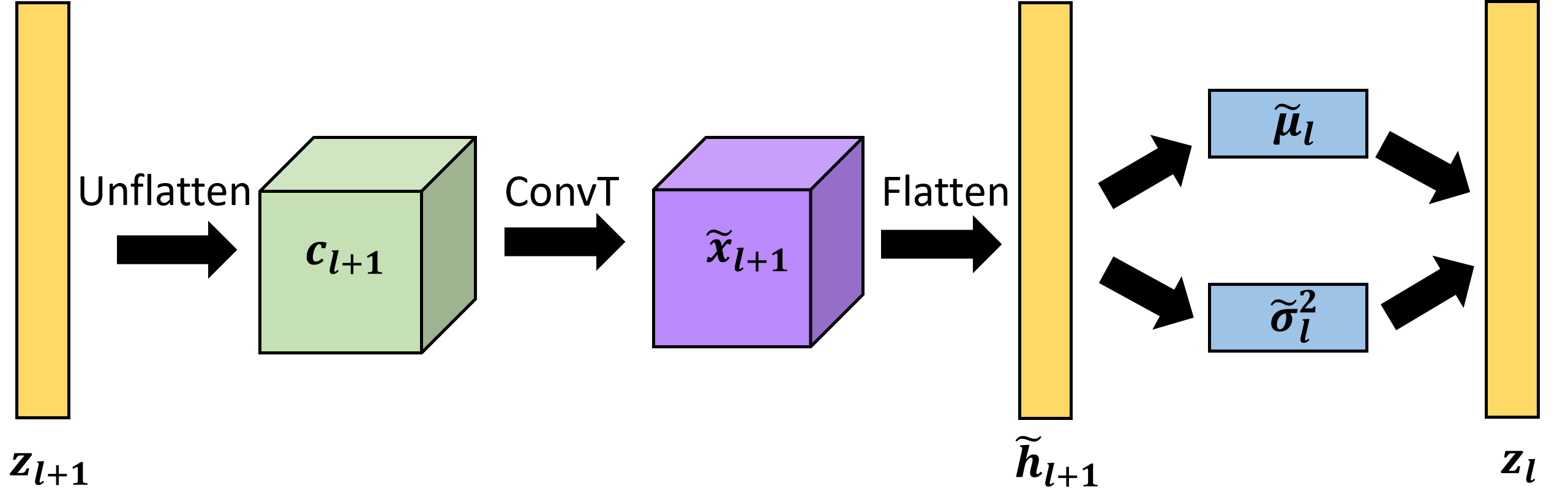}
    \caption{Operations in the downward pathway of CPGM-VAE.}
    \label{fig:o2}
\end{figure}

\textbf{Decoder} $\bm{{\mathcal{G}}.}$ The $l$-th layer in the decoder ${\mathcal{G}}$ is expressed as follows:

\begin{equation*}
\begin{aligned}
&\bm{\tilde c}_{l+1}=\text{Unflatten}(\bm{\tilde z}_{l+1})\\
&\bm{\tilde x}_{l+1}=\text{ConvT}(\bm{\tilde c}_{l+1})\\
&\bm{\tilde h}_{l+1}=\text{Flatten}(\bm{\tilde x}_{l+1})\\
&\bm{\tilde \mu}_l=\text{Linear}(\bm{\tilde h}_{l+1})\\
&\bm{\tilde \sigma}^{2}_l=\text{Softplus}(\text{Linear}(\bm{\tilde h}_{l+1})\\
&\bm{z}_l=\bm{\tilde \mu}_l+\bm{\tilde \sigma}^{2}_l\odot \bm{\epsilon}
\end{aligned}
\end{equation*} 
where \verb ConvT  is a transposed convolutional layer and \verb Unflatten  is a linear layer to convert 1-dimensional data into 2-dimension (Fig.~\ref{fig:o2} illustrates these operations). In the $l$-th layer, the bottom-up information ($\bm{\mu}_l$ and $\bm{\sigma}_l$) and top-down information ($\bm{\tilde \mu}_l$ and $\bm{\tilde \sigma}_l$) are interacted by the following equations defined in \cite{ladder}:
\begin{equation}
\bm{q\_\mu}_{l}=\frac{\bm{\tilde \mu}_l\bm{\tilde \sigma}^{-2}_{l}+\bm{\mu}_l\bm{\sigma}^{-2}_{l}}{\bm{\tilde \sigma}^{-2}_{l}+\bm{\sigma}^{-2}_{l}}
\end{equation}
\begin{equation}
\bm{q\_\sigma^2}_{l}=\frac{1}{\bm{\tilde \sigma}^{-2}_{l}+\bm{\sigma}^{-2}_{l}}
\end{equation}

\textbf{Known Classifier} $\bm{{\mathcal{C}}.}$ The known classifier ${\mathcal{C}}$ is a Softmax layer, which takes the latent representation $\bm{z}$ as input. It produces the probability distribution over the known classes.

\textbf{Unknown Detector} $\bm{{\mathcal{D}}.}$ When training is completed, the unknown detector ${\mathcal{D}}$ is modeled by information hidden in the latent representations and reconstruction errors. During the testing phase, the unknown detector ${\mathcal{D}}$ is used as a binary classifier to judge whether the input is known or unknown (details are discussed in Sec.~\ref{vae_test}).

\subsubsection{Training} \label{vae_train}
During the training phase, the proposed method CPGM-VAE forces the conditional posterior distributions $q_{\bm{\phi}}(\bm{z}|\bm{x},k)$ to approximate different multivariate Gaussian models $p_{\bm{\theta}}^{(k)}(\bm{z})={\mathcal{N}}(\bm{z};\bm{\mu}_{k},\textbf{I})$ where $k$ is the index of known classes, and the mean of $k$-th Gaussian distribution $\bm{\mu}_k$ is obtained by a fully-connected layer which maps the one-hot encoding of the input's label to the latent space. The KL-divergence in latent space (Eqn.~\ref{kl_vae}) is modified as follows:

\begin{equation}
\begin{aligned}
&\minus D_{KL}(q_{\bm{\phi}}(\bm{z}|\bm{x},k) \ || \ p_{\bm{\theta}}^{(k)}(\bm{z}))\\
&=\int q_{\bm{\phi}}(\bm{z}|\bm{x},k)\big(\log p_{\bm{\theta}}^{(k)}(\bm{z})-\log q_{\bm{\phi}}(\bm{z}|\bm{x},k)\big)d\bm{z}\\
&=\int {\mathcal{N}}(\bm{z};\bm{\mu},\bm{\sigma}^2)\big(\log {\mathcal{N}}(\bm{z};\bm{\mu}_{k},\textbf{I})-\log {\mathcal{N}}(\bm{z};\bm{\mu},\bm{\sigma}^2)\big)d\bm{z}\\
&=\frac{1}{2}\sum_{j=1}^{J}\big(1+\log(\sigma_{j}^{2})-(\mu_j-\mu_j^{(k)})^2-\sigma_{j}^{2}\big)
\end{aligned}
\end{equation}

During the training phase, the model is trained to minimize the sum of the reconstruction loss ${\mathcal{L}_r}$, KL-divergence ${\mathcal{L}_{KL}}$ and classification loss ${\mathcal{L}_c}$. To measure classification loss ${\mathcal{L}_c}$, we use softmax cross-entropy of prediction and ground-truth labels. The classification loss is defined as
\begin{equation}
{\mathcal{L}_c}=-\ln S_c
\end{equation}
where $S_c$ is the probability of the target class. To measure reconstruction loss ${\mathcal{L}_r}$, we use the ${{L}_2}$ distance between input image $\bm{x}$ and reconstructed image $\bm{\tilde x}$. The reconstruction loss ${\mathcal{L}_r}$ is defined as
\begin{equation}
{\mathcal{L}_r}=||\bm{x}-\bm{\tilde x}||^2_2
\end{equation}

As the  probabilistic ladder architecture is adopted, the KL-divergence is considered not only in the latent space but also in all middle layers:

\begin{equation}
\begin{aligned}
{\mathcal{L}_{KL}} = &\minus\frac{1}{L}\big[ D_{KL}\big(q_{\bm{\phi}}(\bm{z}|\bm{x},k) \ || \ p_{\bm{\theta}}^{(k)}(\bm{z})\big)\\
&+\sum_{l=1}^{L-1}D_{KL}\big(q_{\bm{\theta}}(\bm{\tilde x}_l|\bm{\tilde x}_{l+1}, \bm{x}) \ || \ q_{\bm{\theta}}(\bm{\tilde x}_l|\bm{\tilde x}_{l+1})\big)\big]
\label{kl}
\end{aligned}
\end{equation}
where 
\begin{equation}
q_{\bm{\theta}}(\bm{\tilde x}_l|\bm{\tilde x}_{l+1}, \bm{x})={\mathcal{N}}(\bm{\tilde x}_l;\bm{q\_\mu}_{l},\bm{q\_\sigma^2}_{l}) 
\end{equation}
\begin{equation}
q_{\bm{\theta}}(\bm{\tilde x}_l|\bm{\tilde x}_{l+1})={\mathcal{N}}(\bm{\tilde x}_l;\bm{\tilde \mu}_{l},\bm{\tilde \sigma^2}_{l})
\end{equation}

The loss function used in our model is summarized as follows:
\begin{equation}
{\mathcal{L}}=\minus({\mathcal{L}_r}+\beta{\mathcal{L}_{KL}}+\lambda{\mathcal{L}_{c})}
\label{vae_loss}
\end{equation}
where $\beta$ is increased linearly from 0 to 1 during the training phase as described in \cite{ladder} and $\lambda$ is a constant. 

\subsubsection{Testing}
\label{vae_test}
When training is completed, we use learned features to model per class multivariate Gaussian models $f_{k}(\bm{z})={\mathcal{N}}(\bm{z};\bm{m}_k,\bm{\sigma}_k^2)$ where $\bm{m}_k$ and $\bm{\sigma}_k^2$ are the mean and variance of the latent representations of all correctly classified training samples in $k$-th class. If the dimension of the latent space is $n$: $\bm{z}=(z_1,..., z_n)$, the probability of a sample locating in the distribution $f_{k}(\bm{z})$ is defined as follows:

\begin{equation}
P_k(\bm{z})=1-\int_{m_0-|z_0-m_0|}^{m_0+|z_0-m_0|} \cdots \int_{m_n-|z_n-m_n|}^{m_n+|z_n-m_n|} f_{k}(\bm{t}) \ d\bm{t}
\end{equation}

We also analyze information hidden in the reconstruction errors. The reconstruction errors of input from known classes are commonly smaller than that of unknown classes \cite{c2ae}. Here we obtain the reconstruction error threshold by ensuring 95\% training data to be recognized as known. We define the threshold of conditional Gaussian distributions (CGD) is $\tau_l$ and the threshold of reconstruction errors (RE) is $\tau_r$. The testing phase can be summarized as follows:  for a testing sample $X$, assuming its latent feature is $Z$ and its reconstruction error is $R$, if ${\forall}k\in \{1,...,K\}, P_k(\bm{Z})<\tau_l \ \textbf{or} \ R>\tau_r$, then $X$ will be predicted as unknown, otherwise, $X$ will be predicted as known with label $y_{p}=argmax(\mathcal{C}(Z))$. Details of the testing procedure are described in Algo.~\ref{algo1}. 

\begin{algorithm}
\caption{Testing procedure of CPGM-VAE}\label{alg:euclid}
\begin{algorithmic}[1]
\Require Testing sample $\bm{X}$
\Require Trained modules ${\mathcal{F}}$, ${\mathcal{G}}$, ${\mathcal{C}}$
\Require Threshold $\tau_l$ of CGD
\Require Threshold $\tau_r$ of RE
\Require For each class $k$, let $\bm{z}_{i,k}$ is the latent representation of each correctly classified training sample $\bm{x}_{i,k}$
    \State \textbf{for} $k=1,\ldots,K$ \textbf{do}
    \State \quad compute the mean and variance of each class: $\bm{m}_k=mean_i(\bm{z}_{i,k})$, $\bm{\sigma}_k^2=var_i(\bm{z}_{i,k})$
    \State \quad model the per class multivariate Gaussian: $f_{k}(\bm{z})={\mathcal{N}}(\bm{z};\bm{m}_k,\bm{\sigma}_k^2)$
    \State \textbf{end for}
    \State latent representation $\bm{Z}={\mathcal{F}}(\bm{X})$
    \State predicted known label $y_{pred}=argmax({\mathcal{C}}\big(\bm{Z})\big)$
    \State reconstructed image $\bm{\tilde X}={\mathcal{G}}(\bm{Z})$
    \State reconstruction error $R=||\bm{X}-\tilde{\bm{X}}||^2_2$
    \State \textbf{if} ${\forall}k\in \{1,...,K\}, P_k(\bm{Z})<\tau_l \ \textbf{or} \ R>\tau_r$ \textbf{then}
    \State \quad predict $\bm{X}$ as unknown
    \State \textbf{else}
    \State \quad predict $\bm{X}$ as known with label $y_{pred}$
    \State \textbf{end if}
\end{algorithmic}
\label{algo1}
\end{algorithm}

\subsection{CPGM-AAE}
\subsubsection{Architecture} 
\label{archi_CPGM-AAE}

\begin{figure*}[htb]
    \centering %
\begin{subfigure}{0.28\textwidth}
  \includegraphics[width=\linewidth, height=3.8cm]{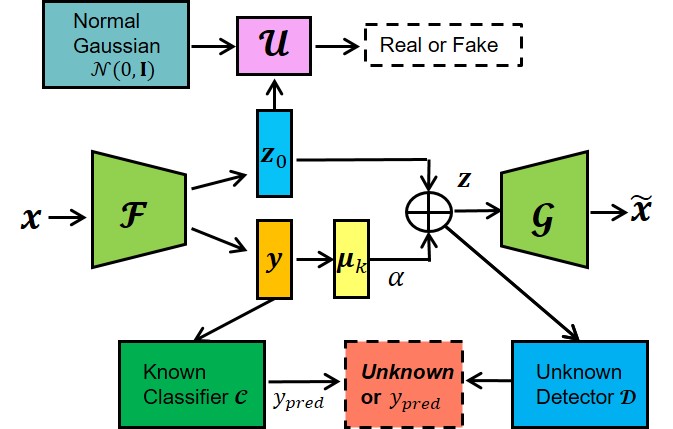}
  \caption{Architecture of CPGM-AAE}
  \label{model3_fra}
\end{subfigure}\hfil %
\begin{subfigure}{0.28\textwidth}
  \includegraphics[width=\linewidth, height=3.8cm]{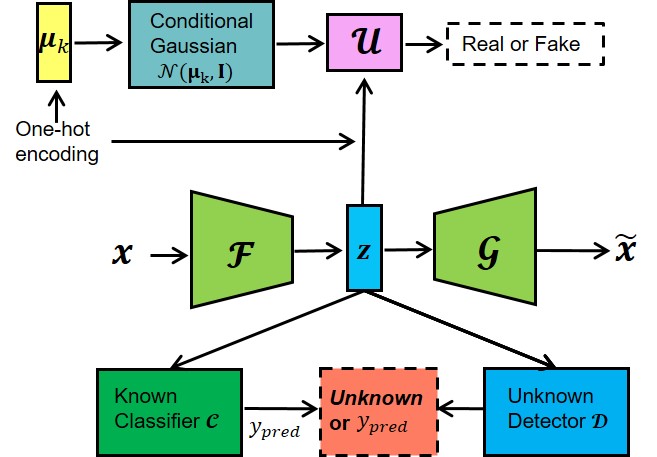}
  \caption{Architecture of AAE-Variant 1}
  \label{model1_fra}
\end{subfigure}\hfil %
\begin{subfigure}{0.28\textwidth}
  \includegraphics[width=\linewidth, height=3.8cm]{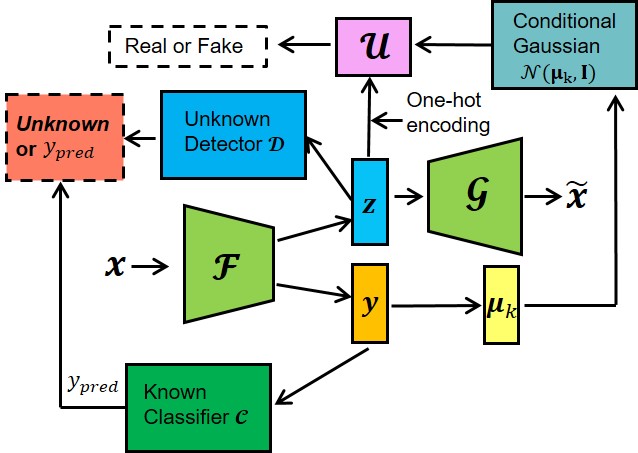}
  \caption{Architecture of AAE-Variant 2}
  \label{model2_fra}
\end{subfigure}
\medskip
\begin{subfigure}{0.28\textwidth}
  \includegraphics[width=\linewidth, height=3.8cm]{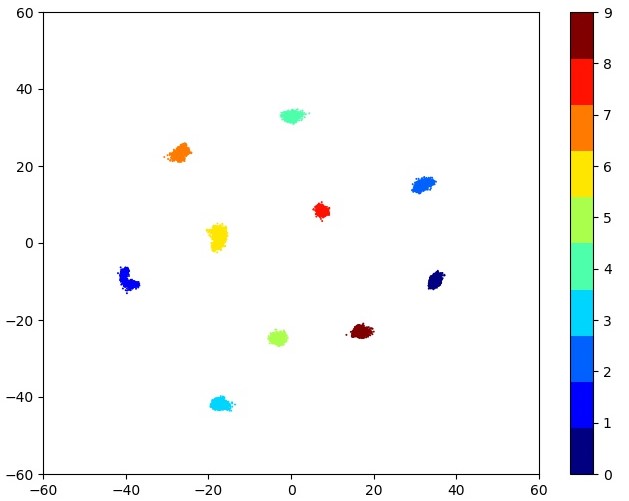}
  \caption{Feature Visualization of CPGM-AAE}
  \label{model3_vis}
\end{subfigure}\hfil %
\begin{subfigure}{0.28\textwidth}
  \includegraphics[width=\linewidth, height=3.8cm]{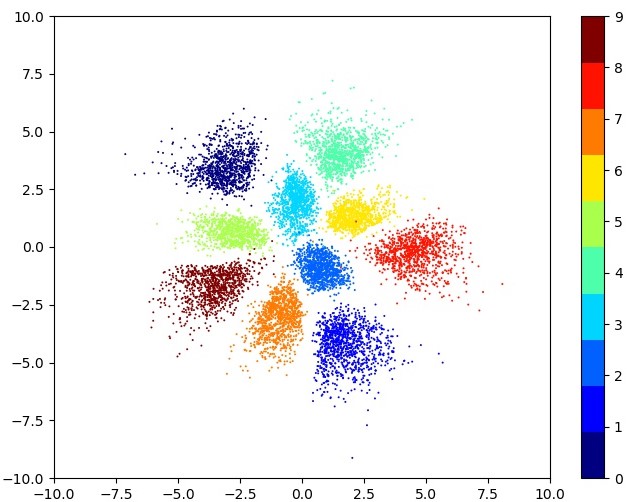}
  \caption{Feature Visualization of AAE-Variant 1}
  \label{model1_vis}
\end{subfigure}\hfil %
\begin{subfigure}{0.28\textwidth}
  \includegraphics[width=\linewidth, height=3.8cm]{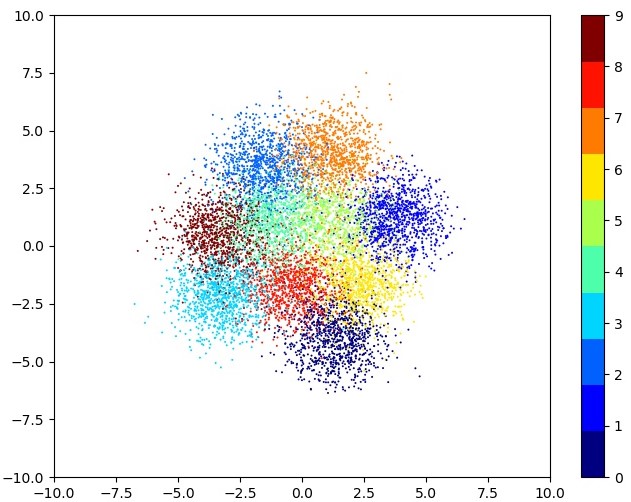}
  \caption{Feature Visualization of AAE-Variant 2}
  \label{model2_vis}
\end{subfigure}
\caption{
The model architecture of the proposed CPGM-AAE \textbf{(a)} (details of CPGM-AAE are described in Sec.~\ref{archi_CPGM-AAE}) and two model variants (\textbf{(b)} and \textbf{(c)}) (details of two model variants are described in Sec.~\ref{model_comp}). We visualize the features of samples from known classes generated by these three models in \textbf{(d)}, \textbf{(e)} and \textbf{(f)}, respectively.
All these three models could generate 10 feature clusters and each cluster represents one class. Although AAE-Variant 1 could generate separable features, these features do not follow class conditional Gaussian distributions for unknown detection. Conversely, AAE-Variant 2 could generate ideal class conditional Gaussian distributions, but they are highly overlapped and can hardly provide discriminative information for known classification. In comparison, the proposed method, CPGM-AAE, could generate ideal class conditional Gaussian distributions for unknown detection, and these features are also separable in the feature space for known classification.}
\label{model_analysis}
\end{figure*}

The architecture of the proposed method CPGM-AAE is shown in Fig.~\ref{model3_fra}, which is composed of five modules:\\
1. Encoder (Generator) ${\mathcal{F}}$\\ 
2. Decoder ${\mathcal{G}}$\\
3. Discriminator ${\mathcal{U}}$\\
4. Known Classifier ${\mathcal{C}}$\\
5. Unknown Detector ${\mathcal{D}}$

To enable AAE to generate discriminative features for known classification, different from traditional VAE models that only extract latent features, we add one more fully-connect layer on the top of the encoder ${\mathcal{F}}(\bm{y},\bm{z_0}|\bm{x})$ to extract categorical information of input. Therefore, the encoder ${\mathcal{F}}(\bm{y},\bm{z_0}|\bm{x})$ provides not only a latent feature $\bm{z_0}$, but also a categorical distribution $\bm{y}$ in the top layer, which can be expressed as follows:

\begin{equation*}
\begin{aligned}
&\bm{h}_n=\text{Flatten}(\bm{x}_n)\\
&\bm{z_0}=\text{Linear}(\bm{h}_n)\\
&\bm{y}=\text{Softmax}(\text{Linear}(\bm{h}_n))\\
\end{aligned}
\end{equation*} 
where \verb Softmax  applies the Softmax function to the categorical distribution $\bm{y}$ so that the elements of $\bm{y}$ lie in the range [0,1] and sum to 1. The latent feature $\bm{z_0}$ will be fed to the discriminator ${\mathcal{U}}$ to approximate a normal Gaussian distribution ${\mathcal{N}}(\bm{z_0};0,\textbf{I})$, while the categorical distribution $\bm{y}$ will be fed to the known classifier ${\mathcal{C}}$ (a Softmax layer) to produce the probability distribution over the known classes for known classification, and a fully-connected layer is used to map $\bm{y}$ into the latent space to generate class conditional Gaussian means $\mu_k$ where $k$ is the index of known classes. By adding $\mu_k$ to the latent feature $\bm{z_0}$: $\bm{z}=\alpha \times \bm{\mu}_{k}+\bm{z_0}$ where $\alpha$ is a constant, different latent features $\bm{z}$ will be close to different means $\mu_k$ to generate class conditional Gaussian distributions ${\mathcal{N}_k}(\bm{z};\bm{\mu}_{k},\textbf{I})$. These learned Gaussian models build up the unknown detector ${\mathcal{D}}$, which will be used during the testing phase to judge whether an input is known or unknown. 

We also explore two model variants that are based on AAE, as shown in Fig.~\ref{model1_fra} and Fig.~\ref{model2_fra}. The main difference between these three models is how to obtain Gaussian means $\mu_k$ to generate conditional Gaussian distributions. In AAE-Variant 1 (as shown in Fig.~\ref{model1_fra}), similar as our previous work (CPGM-VAE), $k$-th class latent features are forced to approximate $k$-th Gaussian distribution ${\mathcal{N}_k}(\bm{z};\bm{\mu}_{k},\textbf{I})$ where Gaussian means $\bm{\mu}_{k}$ are obtained from a separate fully-connected layer that maps the one-hot encoding of the input’s label to the latent space. While in AAE-Variant 2 (as shown in Fig.~\ref{model2_fra}), Gaussian means $\bm{\mu}_{k}$ are achieved by mapping the learned categorical distribution $\bm{y}$ into the latent space through a fully-connected layer. Details of comparison between CPGM-AAE, AAE-Variant 1 and AAE-Variant 2 are discussed in Sec.~\ref{model_comp}.

\subsubsection{Training} \label{aae_train}
There are four phases during the CPGM-AAE training procedure: the \emph{reconstruction} phase, the \emph{regularization} phase, the \emph{classification} phase and the \emph{center-learning} phase. 

In the \emph{reconstruction} phase, the encoder ${\mathcal{F}}$ and decoder ${\mathcal{G}}$ are trained jointly to minimize the reconstruction loss $\mathcal{L}_r$: the $L_2$ distance between the input image $\bm{x}$ and the reconstructed image $\bm{\tilde x}$, which is defined as follows:
\begin{equation}
{\mathcal{L}_r}=||\bm{x}-\bm{\tilde x}||^2_2
\end{equation}

In the \emph{regularization} phase, firstly the discriminator ${\mathcal{U}}$ is trained to distinguish the prior distribution $p(\bm{z})$ from the posterior distribution $q(\bm{z}|\bm{x})={\mathcal{F}}(x)$, then the encoder ${\mathcal{F}}$ is trained to confuse the discriminator ${\mathcal{U}}$. This adversarial process can be expressed in the following equation:

\begin{equation}
\mathop{min}\limits_{\mathcal{F}}\mathop{max}\limits_{\mathcal{U}}E_{p(\bm{z})}[\log {\mathcal{U}}(p(\bm{z}))]+E_{\bm{x}\sim p_{data}}[\log(1-{\mathcal{U}}({\mathcal{F}}(\bm{x})))]
\end{equation}

In the \emph{classification} phase, the known classifier ${\mathcal{C}}$ is trained to minimize the classification loss $\mathcal{L}_c$ between predicted scores and ground-truth labels, which is defined as follows:
\begin{equation}
{\mathcal{L}_c}=-\ln S_c
\end{equation}
where $S_c$ is the probability of the target class.

In the \emph{center-learning} phase, the mean of the $k$-th multivariate Gaussian distribution $\bm{\mu}_k$ is achieved by mapping the categorical distribution $\bm{y}$ into the latent space. Meanwhile, to keep the different distributions as far away as possible, we add a distance loss $\mathcal{L}_d$ to linearly penalize the Euclidean distance between every two centers. Specifically, if the Euclidean distance is larger than a threshold $\eta$, the distance loss $\mathcal{L}_d$ will be zero. This distance loss $\mathcal{L}_d$ can be expressed as the following equation:
\begin{equation}
\mathcal{L}_d(i,j)=\left\{
\begin{array}{rcl}
&||\bm{\mu}_i-\bm{\mu}_j||^2_2 & {||\bm{\mu}_i-\bm{\mu}_j||^2_2 \leq \eta}\\
&0 & {||\bm{\mu}_i-\bm{\mu}_j||^2_2 > \eta}
\end{array} \right. 
\label{eudis}
\end{equation}

\subsubsection{Testing} 
As latent features are also forced to approximate class conditional Gaussian distributions, the testing phase of CPGM-AAE is the same as CPGM-VAE, which has been described in Sec~\ref{vae_test} and summarized in Algo.~\ref{algo1}.
\section{Experiments}
To evaluate the performance of the proposed methods for OSR, we conduct extensive experiments on multiple standard image datasets. In this section, firstly we present important implementation details and introduce used datasets in our experiments. Then, we conduct various ablative experiments to validate each component in the proposed CPGM-VAE and CPGM-AAE. We also compare CPGM-AAE with two AAE-based variants from qualitative analysis and quantitative analysis. Then we compare proposed methods with state-of-the-art methods on popular benchmark datasets. Finally, we conduct a case study to investigate and compare the characteristics between the proposed CPGM-VAE and CPGM-AAE for OSR.

\subsection{Implementation Details}
For a fair comparison with previous works, we employ a re-designed VGGNet defined in \cite{crosr} as our model backbone, which is widely used in the open set recognition literature. The batch size is fixed to 64 and the dimensionality of the latent representation $\bm{z}$ is fixed to 32. The networks were trained without any large degradation in closed set accuracy from the original ones. The closed set accuracies of the networks for each dataset are listed in Table.~\ref{close}. In OSR experiments, the threshold $\tau_r$ of reconstruction errors is obtained by ensuring 95\% training data be recognized as known, and the threshold of conditional Gaussian distributions $\tau_l$ is set to 0.5. 

Especially, for CPGM-VAE, we use the SGD optimizer with a learning rate of 0.001, and the parameter $\beta$ in the loss function (Eqn.~\ref{vae_loss}) is increased linearly from 0 to 1 during the training phase as described in \cite{ladder}, while the parameter $\lambda$ in Eqn.~\ref{vae_loss} is set equal to 100. For CPGM-AAE, we use the SGD optimizer with a learning rate of 0.1. The constant $\alpha$ (described in Sec.~\ref{archi_CPGM-AAE}) and the Euclidean distance threshold $\eta$ in Eqn.~\ref{eudis} are set equal to 10 and 1, respectively. The dimensionality of the categorical distribution $\bm{y}$ equals to the number of known classes.

\begin{table}
\begin{center}
\begin{tabular}{l c c c c}
\hline
Architecture  & MNIST & SVHN & CIFAR-10\\
\hline\hline
Plain CNN & 0.997 & 0.944 & 0.912\\
\hline
CPGM-VAE & 0.996 & 0.942 & 0.912\\
CPGM-AAE & 0.995 & 0.937 & 0.906\\
\hline
\end{tabular}
\end{center}
\caption{Comparison of closed set testing accuracies between the plain CNN and the proposed CPGM methods. Although the training objective of CPGM is classifying known samples as well as learning conditional Gaussian distributions, there is no significant degradation in closed set accuracy.}
\label{close}
\end{table}

\subsection{Dataset Description}
We conduct OSR experiments on six popular benchmark datasets: MNIST \cite{mnist}, SVHN \cite{svhn}, CIFAR-10 \cite{cifar10}, CIFAR-100 \cite{cifar100}, ImageNet \cite{imagenet}, and LSUN \cite{lsun}.

\textbf{MNIST.} MNIST \cite{mnist} is the most popular hand-written digit benchmark. There are 60,000 images for training and 10,000 images for testing from ten digit classes. Each image is monochrome with a size of 28$\times$28. Although near 100\% accuracy has been achieved in the closed set classification \cite{mnist_acc}, the open set extension of MNIST still remains a challenge due to the variety of possible outliers.

\textbf{SVHN.} The Street View House Numbers (SVHN) \cite{svhn} is a dataset consisting of ten digit classes each with between 9,981 and 11,379 color images with a size of 32$\times$32.

\textbf{CIFAR-10.} The CIFAR-10 dataset \cite{cifar10} consists of 5,000 training images and 1,000 testing images in each of the ten natural classes. Each image is colorful with a size of 32$\times$32. In CIFAR-10, each class has large intra-class diversities by color, style, or pose difference.

\textbf{CIFAR-100.} The CIFAR-100 dataset \cite{cifar100} is just like the CIFAR-10 dataset, except it has 100 natural classes containing 600 images each. There are 500 training images and 100 testing images in each class.

\textbf{ImageNet.} ImageNet \cite{imagenet} is a dataset of over 15 million labeled high-resolution images belonging to roughly 22,000 categories. The images were collected from the web and labeled by human labor. 

\textbf{LSUN.} The LSUN dataset \cite{lsun} contains around 10 million labeled images in 10
scene categories (around one million labeled images in each category) and 59 million labeled images in 20 object categories.

\subsection{Ablation Study of CPGM-VAE} \label{ab_vae}
In this section, we analyze our contributions from each component of the proposed CPGM-VAE on the CIFAR-100 dataset~\cite{cifar100}. The CIFAR-100 dataset consists of 100 classes, containing 500 training images and 100 testing images in each class. For ablation analysis, the performance is measured by F-measure (or F1-scores) \cite{f1} against varying Openness \cite{toward}. Openness is defined as follows:

\begin{equation}
    Openness=1-\sqrt{\frac {2\times N_{train}}{N_{test}+N_{target}}}
\end{equation}
where $N_{train}$ is the number of known classes seen during training, $N_{test}$ is the number of classes that will be observed during testing, and $N_{target}$ is the number of classes to be recognized during testing. We randomly sample 15 classes out of 100 classes as known classes and varying the number of unknown classes from 15 to 85, which means Openness is varied from 18\% to 49\%. The performance is evaluated by the macro-average F1-scores in 16 classes (15 known classes and \emph{unknown}).

We compare the following baselines for ablation analysis of CPGM-VAE:\\
\textbf{I. CNN}: In this baseline, only the encoder ${\mathcal{F}}$ (without ladder architecture) and the known classifier ${\mathcal{C}}$ are trained for closed set classification. This model can be viewed as a plain convolutional neural network (CNN). During testing, learned features will be fed to the known classifier ${\mathcal{C}}$ to get the probability scores of known classes. A sample will be recognized as unknown if its maximum probability score of known classes is less than 0.5.\\
\textbf{II. CVAE}: The encoder ${\mathcal{F}}$, decoder ${\mathcal{G}}$ and classifier ${\mathcal{C}}$ are trained without the ladder architecture, and the testing procedure is the same as baseline I. This model can be viewed as a class conditional Variational Autoencoder (CVAE).\\
\textbf{III. LCVAE}: The probabilistic ladder architecture is adopted in the CVAE, which contributes to the KL-divergences during training (Eqn.~\ref{kl}). We call this model as LCVAE. The testing procedure is the same as the baseline I and II.\\
\textbf{IV. CVAE+CGD}: In this baseline, the model architecture and the training procedure are the same as baseline II, while the conditional Gaussian distributions (CGD) are used to detect unknown samples during testing.\\
\textbf{V. LCVAE+CGD}: In this baseline, LCVAE is introduced along with a CGD-based unknown detector. The training and testing procedure are respectively the same as baseline III and IV.\\
\textbf{VI. LCVAE+RE}: In this baseline, different from baseline V, reconstruction errors (RE), instead of CGD, are used in LCVAE to detect unknown samples.\\
\textbf{VII. CPGM-VAE (Proposed Method)}: In this baseline, the training procedure is the same as baseline V and VI, while during testing, CGD and RE are jointly used for unknown detection.\\

\begin{figure} [htbp]
    \centering
    \includegraphics[scale=0.38]{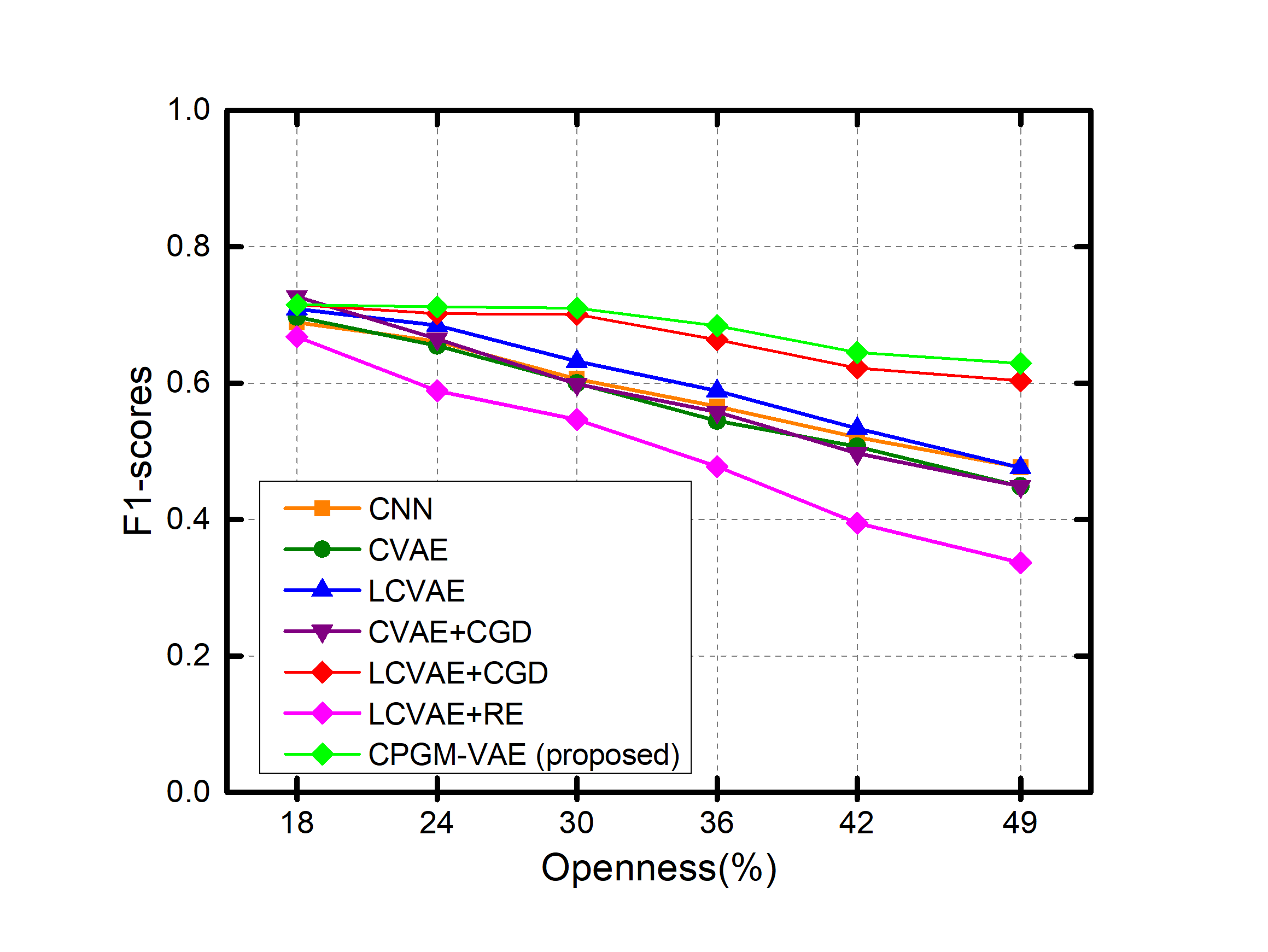}
    \caption{F1-scores against varying Openness with different baselines (described in \ref{ab_vae}) for the ablation study of CPGM-VAE. CPGM-VAE achieves the best F1-scores by using conditional Gaussian distributions (CGD) and reconstruction errors (RE) jointly to detect unknown samples with the probabilistic ladder architecture.}
    \label{abla_vae}
\end{figure}

The experimental results are shown in Fig.~\ref{abla_vae}. Among baseline I (CNN), II (CVAE), and III (LCVAE), unknown detection simply relies on the known classifier ${\mathcal{C}}$, there is no significant improvement in OSR performance. Although the OSR performance is a little improved when the probabilistic ladder architecture is adopted (baseline III: LCVAE), the overall performance in these three baselines is weak as the F1-scores degrade rapidly as the Openness increases. Conditional Gaussian distributions (CGD) are used for unknown detection in the CVAE model (baseline IV: CVAE+CGD), but it has seen no visible change in performance. In baseline V (LCVAE+CGD), this trend is alleviated by introducing a CGD-based unknown detector into LCVAE. This shows the importance of the probabilistic ladder architecture for OSR. It is also the reason why the CGD-based unknown detection achieves better performance in baseline V (LCVAE+CGD) than in baseline IV (CVAE+CGD). If we only use reconstruction errors (RE) to detect unknown samples (baseline VI: LCVAE+RE), the results are worst. However, if RE and CGD are jointly used in the unknown detector (baseline VII: CPGM-VAE), there is a little improvement in performance. As a result, applying conditional Gaussian distributions (CGD) and reconstruction errors (RE) jointly to detect unknown samples with the probabilistic ladder architecture achieves the best performance (baseline VII: CPGM-VAE).
\subsection{Model Analysis of CPGM-AAE} \label{model_comp}
In this section, we compare the characteristics between the proposed CPGM-AAE and two model variants that are based on AAE, as shown in Fig~\ref{model_analysis}. The architectures of these three models are described as follows:

\begin{itemize}
\setlength
    \item CPGM-AAE: The architecture of CPGM-AAE is shown in Fig.~\ref{model3_fra} and has been described in Sec.~\ref{archi_CPGM-AAE}.
    
    \item AAE-Variant 1: The architecture of AAE-Variant 1 is shown in Fig.~\ref{model1_fra}. Similar as our previous work (CPGM-VAE), $k$-th class latent features are forced to approximate $k$-th Gaussian distribution ${\mathcal{N}_k}(\bm{z};\bm{\mu}_{k},\textbf{I})$ in the regularization phase, where Gaussian means $\bm{\mu}_{k}$ are obtained from a separate fully-connected layer that maps the one-hot encoding of the input’s label to the latent space. Inspired from the CGAN model \cite{cgan}, one-hot encoding information is concatenated to latent features, then they are fed into the discriminator ${\mathcal{U}}$ for discriminator training. The latent feature $\bm{z}$ is also fed to the known classifier ${\mathcal{C}}$ for known classification and the unknown detector ${\mathcal{D}}$ for unknown detection, respectively.

    \item AAE-Variant 2: The architecture of AAE-Variant 2 is shown in Fig.~\ref{model2_fra}. Different from AAE-Variant 1, the encoder ${\mathcal{F}}$ is trained not only to obtain the latent feature $\bm{z}$, but also to get the categorical distribution $\bm{y}$. Gaussian means $\bm{\mu}_{k}$ are achieved by mapping the learned categorical distribution $\bm{y}$ into the latent space through a fully-connected layer. Meanwhile, $\bm{y}$ is fed to the known classifier ${\mathcal{C}}$ for known classification, and $\bm{z}$ is fed to the unknown detector ${\mathcal{D}}$ for unknown detection.
\end{itemize}

To evaluate the performance of these three AAE-based models on OSR, we draw 2-dimensional feature visualizations for qualitative analysis and calculate F1-scores against varying Openness \cite{toward} for quantitative analysis.

\textbf{Qualitative Analysis.} For qualitative analysis, we conduct experiments on the MNIST dataset on these three models and draw their latent features on 2-dimensional space respectively for visualization, as shown in Fig.~\ref{model3_vis}, Fig.~\ref{model1_vis}, and Fig.~\ref{model2_vis}. Through these 2-dimensional latent features, we find all these three models could generate 10 feature clusters and each cluster represents one class. Although AAE-Variant 1 could generate separable features, these features do not follow class conditional Gaussian distributions for unknown detection. Conversely, AAE-Variant 2 could generate ideal class conditional Gaussian distributions, but they are highly overlapped and can hardly provide discriminative information for known classification. In comparison, the proposed method, CPGM-AAE, could generate ideal class conditional Gaussian distributions for unknown detection, and these features are also separable in the feature space for known classification.

\begin{figure} [htbp]
  \centering
  \includegraphics[scale=0.35]{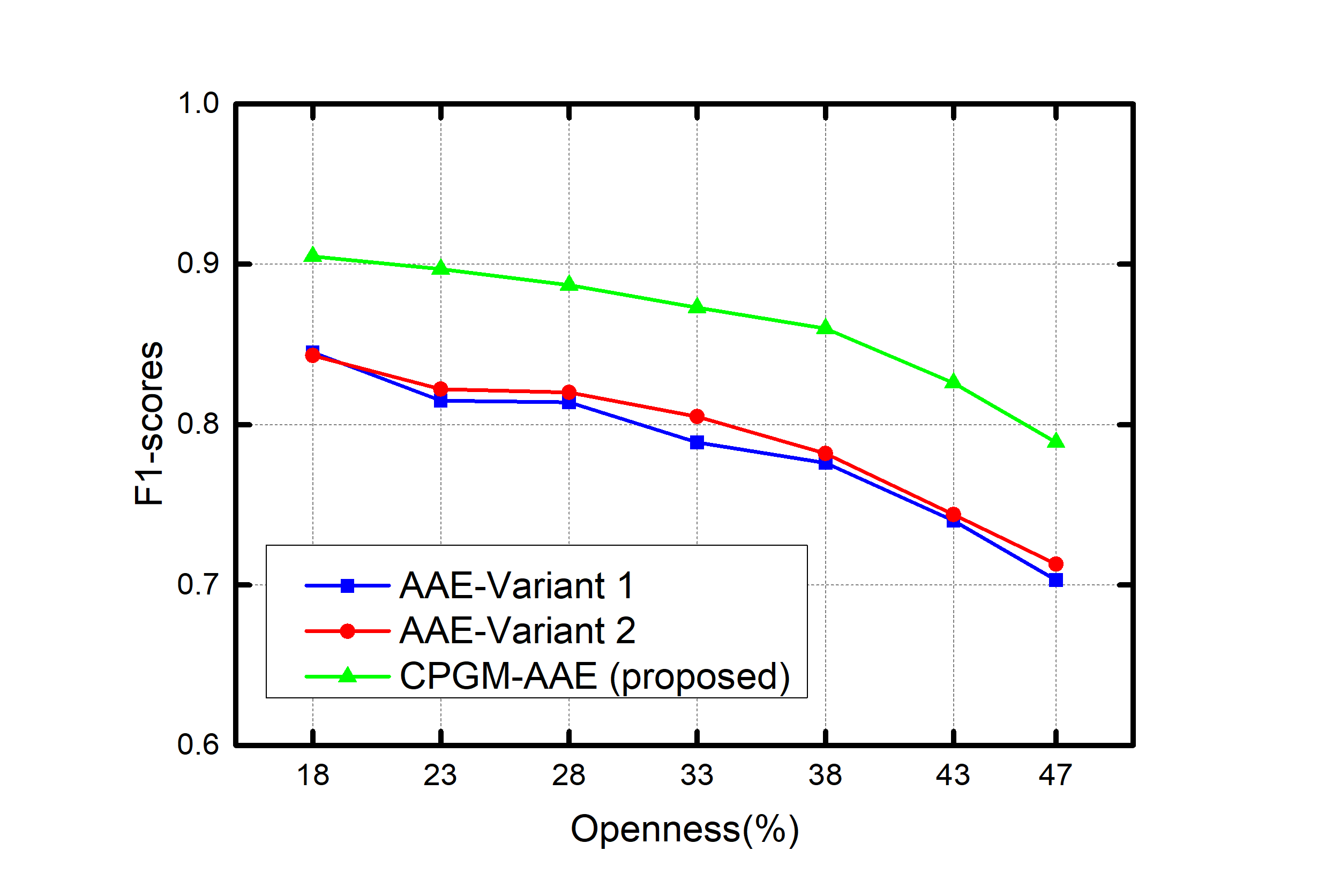}%
  \caption{F1-scores against varying Openness with three AAE-based models (described in \ref{model_comp}). Among these three models,  CPGM-AAE always achieves the best F1-scores under varying Openness.}
  \label{model_aae}
\end{figure}

\textbf{Quantitative Analysis}. We also conduct experiments on these three models for quantitative analysis by calculating F1-scores against varying Openness. Here we still use the MNIST dataset (10 digit classes) as known classes, and use 52 letter classes in the EMNIST dataset \cite{emnist} as unknown classes. Performance is evaluated by macro-average F1-scores \cite{f1} against varying Openness \cite{toward} from 18\% to 47\% by randomly sampling 10, 14, 19, 25, 32, 42, 52 classes respectively from unknown classes. The experimental results are shown in Fig.~\ref{model_aae}. As the Openness increases, F1-scores of these three models all decrease, but CPGM-AAE always achieves the best performance, which is consistent with the results in qualitative analysis.

Moreover, we conduct ablative experiments to explore contributions from each component of CPGM-AAE. We compare the following baselines:\\
\textbf{I. CNN}: In this baseline, only the encoder (generator) ${\mathcal{F}}$ and the known classifier ${\mathcal{C}}$ are trained for closed set classification. This model can be viewed as a plain convolutional neural network (CNN). During testing, learned features will be fed to the known classifier ${\mathcal{C}}$ to get the probability scores of known classes. A sample will be recognized as unknown if its maximum probability score of known classes is less than 0.5.\\
\textbf{II. CAAE}: The encoder (generator) ${\mathcal{F}}$, decoder ${\mathcal{G}}$, discriminator ${\mathcal{U}}$ and known classifier ${\mathcal{C}}$ are trained for closed set classification. The testing procedure is the same as baseline I. This model can be viewed as a class conditional Adversarial Autoencoder (CAAE).\\
\textbf{III. CAAE+CGD}: In this baseline, the model architecture and the training procedure are the same as baseline II, while the conditional Gaussian distributions (CGD) are used to detect unknown samples during testing.\\
\textbf{IV. CAAE+RE}: Different from baseline III, reconstruction errors (RE), instead of CGD, are used in CAAE to detect unknown samples.\\
\textbf{V. CPGM-AAE (Proposed Method)}: In this baseline, the training procedure is the same as baseline III and IV, while during testing, CGD and RE are used jointly for unknown detection.\\

\begin{figure} [htbp]
  \centering
  \includegraphics[scale=0.35]{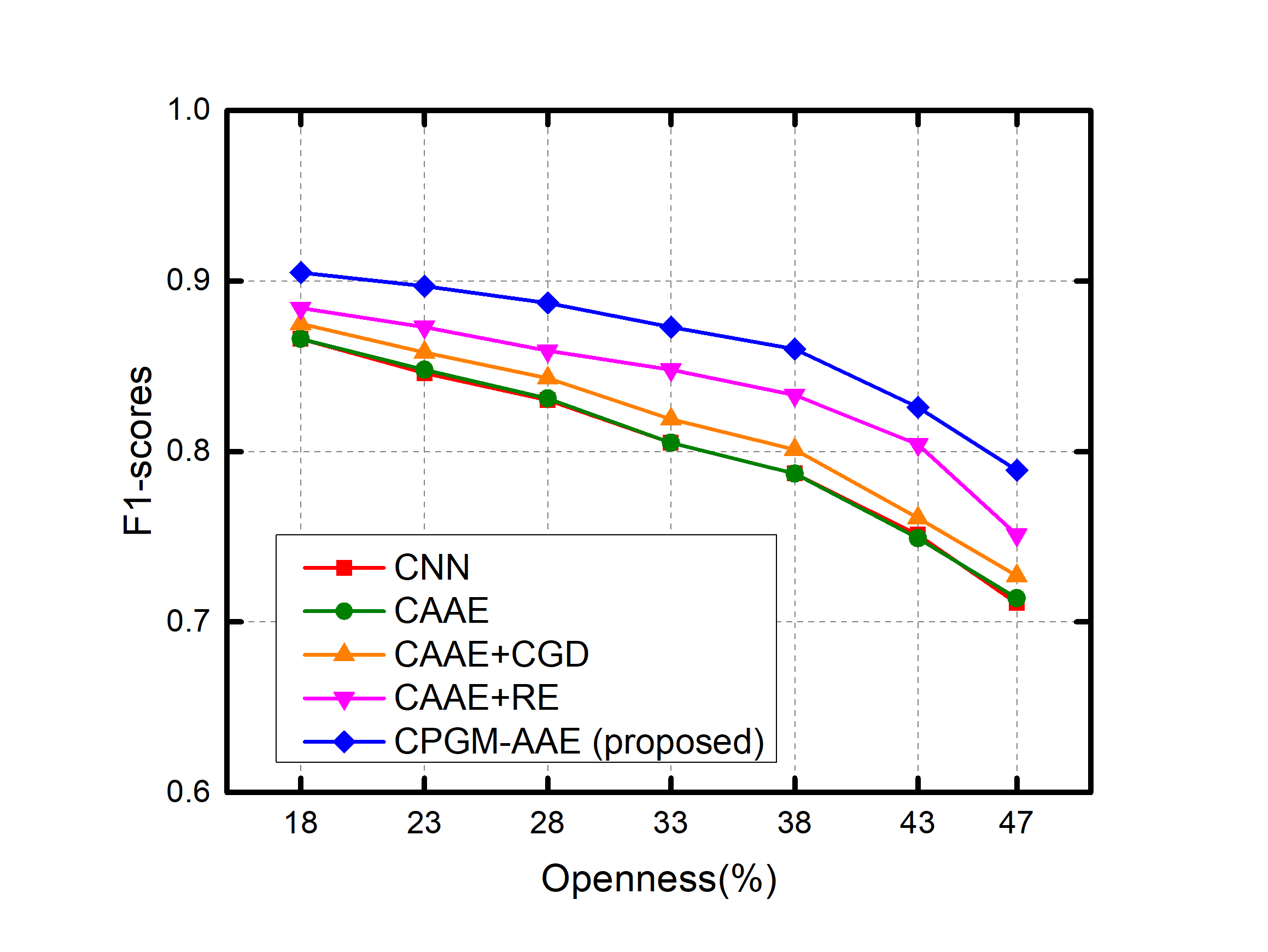}%
  \caption{F1-scores against varying Openness with different baselines (described in \ref{model_comp}) for the ablation study of CPGM-AAE. CPGM-AAE achieves the best F1-scores by using conditional Gaussian distributions (CGD) and reconstruction errors (RE) jointly to detect unknown samples.}
  \label{abla_aae}
\end{figure}

The experimental results are shown in Fig.~\ref{abla_aae}. From the baseline I (CNN) and II (CAAE), unknown detection simply relies on the known classifier ${\mathcal{C}}$, there is no obvious change in OSR performance when adding the reconstruction loss function (baseline II: CAAE) into the training procedure of a plain CNN (baseline I: CNN). However, the performance is a little improved when condition Gaussian distributions (CGD) are used for unknown detection (baseline III: CAAE+CGD). If we only use reconstruction errors (RE) to detect unknown samples (baseline IV: CAAE+RE), the performance has been greatly improved, which has exactly the reverse effect in CPGM-VAE as shown by `LCVAE+RE' in Fig.~\ref{abla_vae} (more analysis will be discussed in Sec.~\ref{case_study}). Baseline V (CPGM-AAE) achieves the best performance where RE and CGD are used jointly for unknown detection.
\subsection{Comparison with State-of-the-art Results}

\begin{table}
\begin{center}
\begin{tabular}{l c c c c }
\hline
Method & Omniglot & MNIST-noise & Noise \\
\hline\hline
Softmax & 0.595 & 0.801 & 0.829\\
Openmax \cite{openmax} & 0.780 & 0.816 & 0.826\\
CROSR \cite{crosr} & 0.793 & 0.827 & 0.826\\
\hline
ours:CPGM-VAE & \textbf{0.850} & \textbf{0.887} & \textbf{0.859}\\
ours:CPGM-AAE & \textbf{0.872} & \textbf{0.865} & \textbf{0.872}\\
\hline
\end{tabular}
\end{center}
\caption{Open set classification results on the MNIST dataset with various outliers added to the test set as unknowns. The performance is evaluated by macro-averaged F1-scores in 11 classes (10 known classes and \emph{unknown}). Our methods significantly outperform the baseline methods.}
\label{tab:state2}
\end{table}

In this section, we compare the proposed method with state-of-the-art methods on OSR tasks. Performance is measured by macro-averaged $F_1$-scores in all known classes and \emph{unknown}. In the following experiments, the models are trained by all known classes, but in the testing phase, samples from unknown classes are added to the test set.

\begin{figure} [htbp]
    \centering
    \includegraphics[scale=0.4]{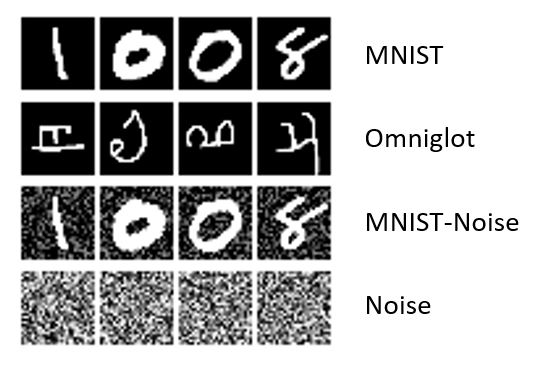}
    \caption{Examples from MNIST, Omniglot, MNIST-Noise, and Noise datasets.}
    \label{fig:out}
\end{figure}

Firstly, we choose MNIST, the most popular hand-written digit dataset, as the training set. As outliers, we follow the set up in \cite{crosr}, using Omniglot \cite{omniglot}, MNIST-Noise, and Noise these three datasets. Omniglot is a dataset containing various alphabet characters. Noise is a synthesized dataset by setting each pixel value independently from a uniform distribution on $[0, 1]$. MNIST-Noise is also a synthesized dataset by adding noise on MNIST testing samples. Each dataset contains 10, 000 testing samples, the same as MNIST, and this makes the known-to-unknown ratio 1:1. Fig.~\ref{fig:out} shows examples of these datasets. The open set recognition scores are shown in Table.~\ref{tab:state2} and the proposed method achieves the best results on all given datasets. It also should be noted that CPGM-AAE is more effective on Omniglot and Noise these two datasets, but not as good as CPGM-VAE on the MNIST-Noise dataset. 

Secondly, following the protocol defined in \cite{crosr}, all samples in the CIFAR-10 dataset are collected as known data, and samples from other datasets, i.e., ImageNet \cite{imagenet} and LSUN \cite{lsun}, are selected as unknown samples. We resize or crop the unknown samples to make them have the same size with known samples. ImageNet-crop, ImageNet-resize, LSUN-crop, and LSUN-resize these four datasets are generated, and each dataset contains 10,000 testing images through random selection, which is the same as CIFAR-10. This makes during testing the known-to-unknown ratio is 1:1. The performance is evaluated by macro-averaged F1-scores in 11 classes (10 known classes and \emph{unknown}), and results are shown in Table.~\ref{tab:state3}. We can see from the results that on all given datasets, the proposed method is more effective than previous methods and achieves a new state-of-the-art performance. It should be noted that CPGM-VAE yields better OSR performance compared to CPGM-AAE on the ImageNet dataset. Conversely, CPGM-AAE is more effective than CPGM-VAE on the LSUN dataset.

\begin{table*}
\begin{center}
\begin{tabular}{l c c c c}
\hline
Method & ImageNet-crop & ImageNet-resize & LSUN-crop & LSUN-resize \\
\hline\hline
Softmax \cite{crosr}$^*$ & 0.639 & 0.653 & 0.642 & 0.647 \\
Openmax \cite{openmax} & 0.660 & 0.684 & 0.657 & 0.668 \\
LadderNet+Softmax \cite{crosr} & 0.640 & 0.646 & 0.644 & 0.647 \\
LadderNet+Openmax \cite{crosr} & 0.653 & 0.670 & 0.652 & 0.659 \\
DHRNet+Softmax \cite{crosr} & 0.645 & 0.649 & 0.650 & 0.649 \\
DHRNet+Openmax \cite{crosr} & 0.655 & 0.675 & 0.656 & 0.664 \\
CROSR \cite{crosr} & 0.721 & 0.735 & 0.720 & 0.749\\
C2AE \cite{c2ae} & 0.837 & 0.826 & 0.783 & 0.801\\
\hline
ours: CPGM-VAE & \textbf{0.840} & \textbf{0.832} & \textbf{0.806} & \textbf{0.812}\\
ours: CPGM-AAE & {0.793} & \textbf{0.830} & \textbf{0.820} & \textbf{0.865}\\
\hline
\end{tabular}
\end{center}
\caption{Open set classification results on the CIFAR-10 dataset with various outliers added to the test set as unknowns. The performance is evaluated by macro-averaged F1-scores in 11 classes (10 known classes and \emph{unknown}). Our methods achieve the best performance among baseline methods.\\ $^*$We report the experimental results reproduced in \cite{crosr}.}
\label{tab:state3}
\end{table*}

\subsection{CPGM-VAE \textit{vs.} CPGM-AAE}
\label{case_study}
We conduct a case study on the CIFAR10 dataset where all animal categories (bird, cat, deer, dog, frog, and horse) are selected as known classes, while all vehicle categories (airplane, car, ship, and truck) are considered to be unknown classes. 
We compare the performance between the proposed CPGM-VAE and CPGM-AAE by by the macro-averaged $F_1$-scores in known classes and \emph{unknown}. CPGM-VAE produces an $F_1$-scores of 0.655 and CPGM-AAE produces an $F_1$-scores of 0.697

\begin{figure}
\begin{subfigure}{\columnwidth}
  \centering
  \includegraphics[scale=0.35]{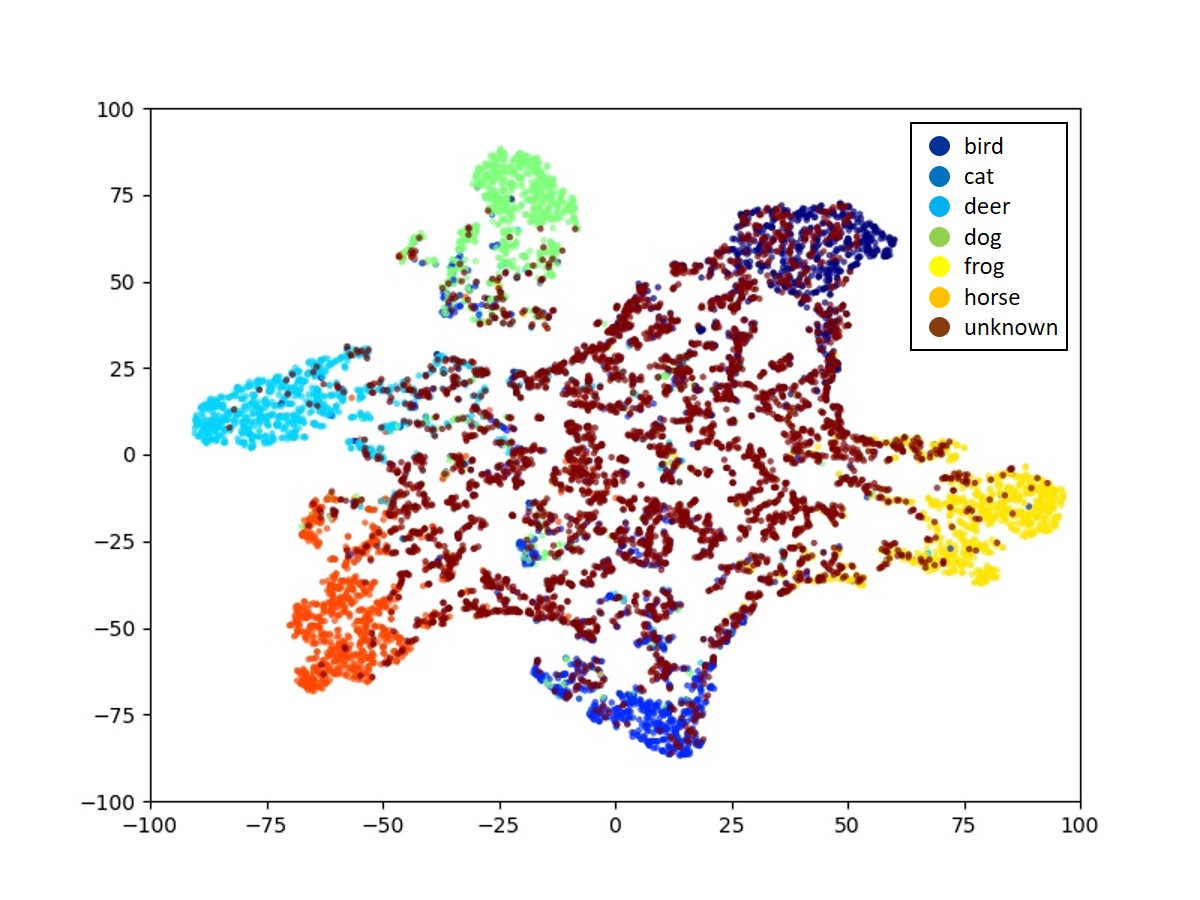}
  \caption{CPGM-VAE}
  \label{vae_case}
\end{subfigure}%
\hspace{.2in}
\begin{subfigure}{\columnwidth}
  \centering
  \includegraphics[scale=0.35]{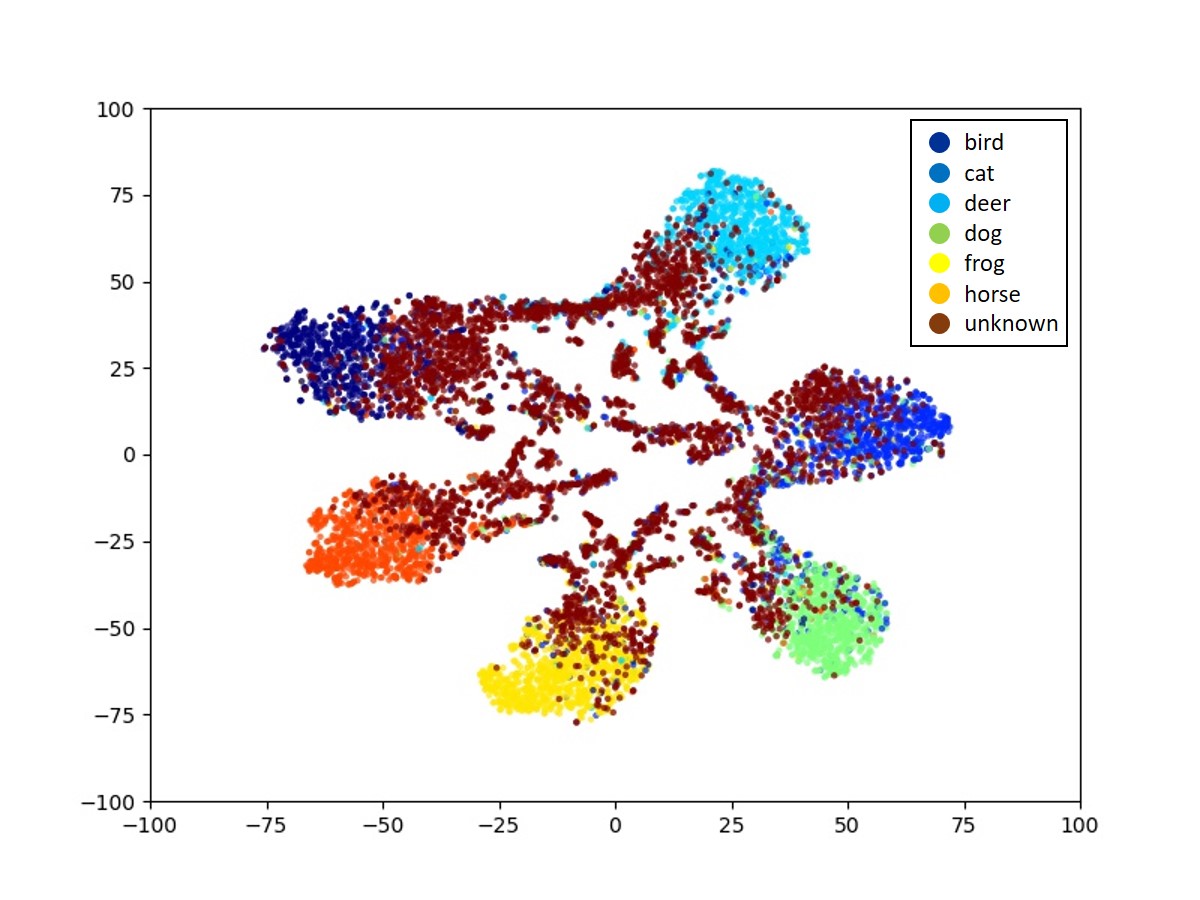}
  \caption{CPGM-AAE}
  \label{aae_case}
\end{subfigure}
\caption{tSNE \cite{tsne} visualization of feature distributions of testing samples in CPGM-VAE \textbf{(a)} and CPGM-AAE \textbf{(b)}. In both methods, samples from different known classes distribute in different clusters while unknown samples are mostly located in open space.
The comparison between CPGM-VAE and CPGM-AAE shows that the variance of each cluster is large in CPGM-VAE and the clusters of known classes are more compact in CPGM-AAE. Therefore, when using conditional Gaussian distributions for OSR, CPGM-AAE yields better results.
}
\label{vis_case}
\end{figure}

To investigate why CPGM-AAE achieves a better performance on OSR, we use tSNE \cite{tsne} to draw the final feature distributions of testing samples on a 2-dimensional space for both CPGM-VAE and CPGM-AAE, as shown in Fig.~\ref{vis_case}. In both cases, known-class features are clustered into six separate clusters and each cluster represents a known class. Most unknown samples are located in open space. However, we note that known clusters appearing under CPGM-AAE (Fig.~\ref{aae_case}) are more compact compared to CPGM-VAE (Fig.~\ref{vae_case}). Therefore, when using conditional Gaussian distributions for OSR, CPGM-AAE yields better results.

\begin{figure}
\begin{subfigure}{0.5\columnwidth}
  \centering
  \includegraphics[scale=0.45]{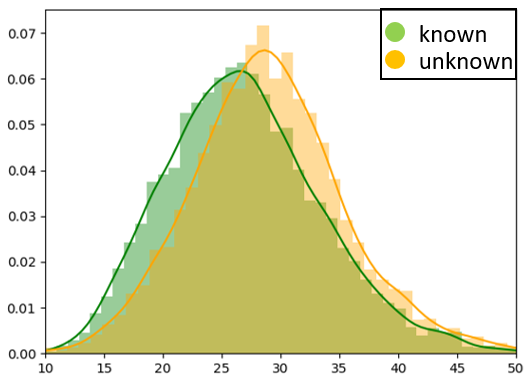}
  \caption{CPGM-VAE}
  \label{vae_rechis}
\end{subfigure}%
\begin{subfigure}{0.5\columnwidth}
  \centering
  \includegraphics[scale=0.45]{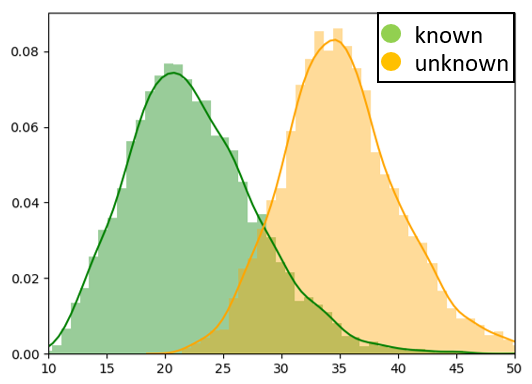}
  \caption{CPGM-AAE}
  \label{aae_rechis}
\end{subfigure}
\caption{Reconstruction error histograms of known and unknown classes in CPGM-VAE \textbf{(a)} and CPGM-AAE \textbf{(b)}. CPGM-AAE has a more clear decision boundary between known and unknown classes, which explains why additionally using reconstruction errors to detect unknown samples at test time can significantly improve performance on CPGM-AAE (the comparison between `CAAE+CGD' and `CPGM-AAE (proposed)' as shown in Fig.~\ref{abla_aae}) but no obvious improvement on CPGM-VAE (the comparison between `LCVAE+CGD' and `CPGM-VAE (proposed)' as shown in Fig.~\ref{abla_vae}).}
\label{rechis}
\end{figure}

\begin{figure*}
  \centering
  \includegraphics[scale=0.65]{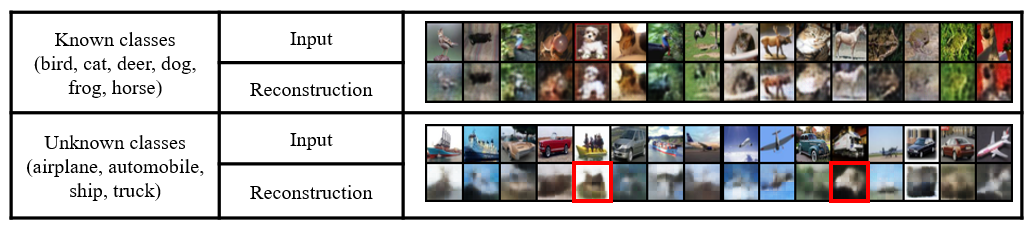}%
  \caption{Reconstructed images generated from CPGM-AAE trained on known classes. All reconstructed images take the form of a blurry version of the input images, but details are preserved better in reconstructions of known samples. Predicting labels from reconstructions of unknown samples is difficult. This model seems to generate a `cat' from a ship and generate a `dog' from a truck (marked in red borders).}
  \label{re_im}
\end{figure*}

Fig.~\ref{rechis} visualizes reconstruction error histograms generated from known (shown in green) and unknown (shown in yellow) samples in CPGM-VAE and CPGM-AAE. As evident from Fig.~\ref{rechis}, CPGM-AAE has a more clear decision boundary between known and unknown classes than CPGM-VAE. This explains why at the test time, additional using reconstruction errors can significantly improve CPGM-AAE performance (the comparison between `CAAE+CGD' and `CPGM-AAE (proposed)' as shown in Fig.~\ref{abla_aae}), while in CPGM-VAE, additional using reconstruction errors only achieves a little improvement (the comparison between `LCVAE+CGD' and `CPGM-VAE (proposed)' as shown in Fig.~\ref{abla_vae}). A possible reason of why CPGM-AAE could generate a more clear decision boundary on reconstruction errors between known and unknown classes is that, like most GAN-based models, the training procedure of CPGM-AAE is divided into several phases and one of these phases is dedicated to reducing reconstruction errors. However, like most VAE-based models, CPGM-VAE is trained not only to reconstruct the input accurately but also to reduce the KL-divergence loss and classification loss at the same phase, which may weaken the ability of VAE models to reconstruct inputs accurately. In Fig.~\ref{re_im}, we visualize reconstructed images (of randomly chosen samples) obtained through CPGM-AAE.
According to Fig.~\ref{re_im}, all reconstructed images take the form of a blurry version of the input images. However, it should be noted that
reconstructions of known samples carry more details compared to reconstructions of unknown samples. Moreover, it is hard to predict the label of an unknown sample by merely looking at its reconstructed image. For example, this model seems to generate a `cat' from a ship and generate a `dog' from a truck (marked in red borders in Fig.~\ref{re_im}).

\section{Conclusion}
In this paper, we have presented a novel framework, Conditional Probabilistic Generative Models (CPGM), for open set recognition (OSR). We focus on two probabilistic generative models, namely VAE and AAE. Compared with previous methods solely based on VAEs or AAEs, the proposed method can not only classify known samples but also detect unknown samples by forcing posterior distributions in the latent space to approximate different Gaussian models. For CPGM-VAE, the probabilistic ladder architecture is adopted to preserve the information that may vanish in the middle layers. This ladder architecture obviously improves the OSR performance. For CPGM-AAE, three model variants of AAE are explored from qualitative analysis and quantitative analysis to investigate their characteristics for OSR. Experiments on several standard image datasets show that the proposed methods significantly outperform the baseline methods and achieve new state-of-the-art results. A case study between CPGM-VAE and CPGM-AAE is also conducted to investigate and compare their characteristics: CPGM-AAE could generate more compact known clusters for OSR and have a more clear decision boundary between known and unknown classes.

\ifCLASSOPTIONcompsoc
  \section*{Acknowledgments}
\else
  \section*{Acknowledgment}
\fi

This research work was conducted in the SMRT-NTU Smart Urban Rail Corporate Laboratory with funding support from the National Research Foundation (NRF), SMRT and Nanyang Technological University; under the Corp Lab@University Scheme.

\ifCLASSOPTIONcaptionsoff
  \newpage
\fi

\bibliographystyle{IEEEtran}
\bibliography{egbib}

\end{document}